# Join-Graph Propagation Algorithms

**Robert Mateescu**                                                    ROMATEES@MICROSOFT.COM
*Microsoft Research*
*7 J J Thomson Avenue*
*Cambridge CB3 0FB, UK*

**Kalev Kask**                                                             KKASK@ICS.UCI.EDU
*Donald Bren School of Information and Computer Science*
*University of California Irvine*
*Irvine, CA 92697, USA*

**Vibhav Gogate**                                             VGOGATE@CS.WASHINGTON.EDU
*Computer Science & Engineering*
*University of Washington, Seattle*
*Seattle, WA 98195, USA*

**Rina Dechter**                                                        DECHTER@ICS.UCI.EDU
*Donald Bren School of Information and Computer Science*
*University of California Irvine*
*Irvine, CA 92697, USA*

## Abstract

The paper investigates parameterized approximate message-passing schemes that are based on bounded inference and are inspired by Pearl's belief propagation algorithm (BP). We start with the bounded inference mini-clustering algorithm and then move to the iterative scheme called Iterative Join-Graph Propagation (IJGP), that combines both iteration and bounded inference. Algorithm IJGP belongs to the class of Generalized Belief Propagation algorithms, a framework that allowed connections with approximate algorithms from statistical physics and is shown empirically to surpass the performance of mini-clustering and belief propagation, as well as a number of other state-of-the-art algorithms on several classes of networks. We also provide insight into the accuracy of iterative BP and IJGP by relating these algorithms to well known classes of constraint propagation schemes.

## 1. Introduction

Probabilistic inference is the principal task in Bayesian networks and is known to be an NP-hard problem (Cooper, 1990; Roth, 1996). Most of the commonly used exact algorithms such as join-tree clustering (Lauritzen & Spiegelhalter, 1988; Jensen, Lauritzen, & Olesen, 1990) or variable-elimination (Dechter, 1996, 1999; Zhang, Qi, & Poole, 1994), and more recently search schemes (Darwiche, 2001; Bacchus, Dalmao, & Pitassi, 2003; Dechter & Mateescu, 2007) exploit the network structure. While significant advances were made in the last decade in exact algorithms, many real-life problems are too big and too hard, especially when their structure is dense, since they are time and space exponential in the *treewidth* of the graph. Approximate algorithms are therefore necessary for many practical problems, although approximation within given error bounds is also NP-hard (Dagum & Luby, 1993; Roth, 1996).

The paper focuses on two classes of approximation algorithms for the task of belief updating. Both are inspired by Pearl's belief propagation algorithm (Pearl, 1988), which is known to be exact for trees. As a distributed algorithm, Pearl's belief propagation can also be applied iteratively to networks that contain cycles, yielding Iterative Belief Propagation (IBP), also known as loopy belief propagation. When the networks contain cycles, IBP is no longer guaranteed to be exact, but in many cases it provides very good approximations upon convergence. Some notable success cases are those of IBP for coding networks (McEliece, MacKay, & Cheng, 1998; McEliece & Yildirim, 2002), and a version of IBP called survey propagation for some classes of satisfiability problems (Mézard, Parisi, & Zecchina, 2002; Braunstein, Mézard, & Zecchina, 2005).

Although the performance of belief propagation is far from being well understood in general, one of the more promising avenues towards characterizing its behavior came from analogies with statistical physics. It was shown by Yedidia, Freeman, and Weiss (2000, 2001) that belief propagation can only converge to a stationary point of an approximate free energy of the system, called Bethe free energy. Moreover, the Bethe approximation is computed over pairs of variables as terms, and is therefore the simplest version of the more general Kikuchi (1951) cluster variational method, which is computed over clusters of variables. This observation inspired the class of Generalized Belief Propagation (GBP) algorithms, that work by passing messages between clusters of variables. As mentioned by Yedidia et al. (2000), there are many GBP algorithms that correspond to the same Kikuchi approximation. A version based on region graphs, called "canonical" by the authors, was presented by Yedidia et al. (2000, 2001, 2005). Our algorithm Iterative Join-Graph Propagation is a member of the GBP class, although it will not be described in the language of region graphs. Our approach is very similar to and was independently developed from that of McEliece and Yildirim (2002). For more information on BP state of the art research see the recent survey by Koller (2010).

We will first present the *mini-clustering* scheme which is an anytime bounded inference scheme that generalizes the mini-bucket idea. It can be viewed as a belief propagation algorithm over a tree obtained by a relaxation of the network's structure (using the technique of variable duplication). We will subsequently present *Iterative Join-Graph Propagation* (IJGP) that sends messages between clusters that are allowed to form a cyclic structure.

Through these two schemes we investigate: (1) the quality of bounded inference as an anytime scheme (using mini-clustering); (2) the virtues of iterating messages in belief propagation type algorithms, and the result of combining bounded inference with iterative message-passing (in IJGP).

In the background section 2, we overview the Tree-Decomposition scheme that forms the basis for the rest of the paper. By relaxing two requirements of the tree-decomposition, that of connectedness (via mini-clustering) and that of tree structure (by allowing cycles in the underlying graph), we combine bounded inference and iterative message-passing with the basic tree-decomposition scheme, as elaborated in subsequent sections.

In Section 3 we present the partitioning-based anytime algorithm called Mini-Clustering (MC), which is a generalization of the Mini-Buckets algorithm (Dechter & Rish, 2003). It is a message-passing algorithm guided by a user adjustable parameter called *i-bound*, offering a flexible tradeoff between accuracy and efficiency in anytime style (in general the higher the i-bound, the better the accuracy). MC algorithm operates on a tree-decomposition, and similar to Pearl's belief propagation algorithm (Pearl, 1988) it converges in two passes, up and down the tree. Our contribution beyond other works in this area (Dechter & Rish, 1997; Dechter, Kask, & Larrosa, 2001) is in: (1) Extending the partition-based approximation for belief updating from mini-buckets to general tree-decompositions, thus allowing the computation of the updated beliefs for all the variables at once.

This extension is similar to the one proposed by Dechter et al. (2001), but replaces optimization with probabilistic inference. (2) Providing empirical evaluation that demonstrates the effectiveness of the idea of tree-decomposition combined with partition-based approximation for belief updating.

Section 4 introduces the Iterative Join-Graph Propagation (IJGP) algorithm. It operates on a general join-graph decomposition that may contain cycles. It also provides a user adjustable *i-bound* parameter that defines the maximum cluster size of the graph (and hence bounds the complexity), therefore it is both anytime and iterative. While the algorithm IBP is typically presented as a generalization of Pearl's Belief Propagation algorithm, we show that IBP can be viewed as IJGP with the smallest i-bound.

We also provide insight into IJGP's behavior in Section 4. Zero-beliefs are variable-value pairs that have zero conditional probability given the evidence. We show that: (1) if a value of a variable is assessed as having zero-belief in any iteration of IJGP, it remains a zero-belief in all subsequent iterations; (2) IJGP converges in a finite number of iterations relative to its set of zero-beliefs; and, most importantly (3) that the set of zero-beliefs decided by any of the iterative belief propagation methods is sound. Namely any zero-belief determined by IJGP corresponds to a true zero conditional probability relative to the given probability distribution expressed by the Bayesian network.

Empirical results on various classes of problems are included in Section 5, shedding light on the performance of IJGP(i). We see that it is often superior, or otherwise comparable, to other state-of-the-art algorithms.

The paper is based in part on earlier conference papers by Dechter, Kask, and Mateescu (2002), Mateescu, Dechter, and Kask (2002) and Dechter and Mateescu (2003).

## 2. Background

In this section we provide background for exact and approximate probabilistic inference algorithms that form the basis of our work. While we present our algorithms in the context of directed probabilistic networks, they are applicable to any graphical model, including Markov networks.

### 2.1 Preliminaries

**Notations:** A reasoning problem is defined in terms of a set of variables taking values on finite domains and a set of functions defined over these variables. We denote variables or subsets of variables by uppercase letters (e.g., $X, Y, Z, S, R \ldots$) and values of variables by lower case letters (e.g., $x, y, z, s$). An assignment $(X_1 = x_1, \ldots, X_n = x_n)$ can be abbreviated as $x = (x_1, \ldots, x_n)$. For a subset of variables $S$, $D_S$ denotes the Cartesian product of the domains of variables in $S$. $x_S$ is the projection of $x = (x_1, \ldots, x_n)$ over a subset $S$. We denote functions by letters $f$, $g$, $h$, etc., and the scope (set of arguments) of the function $f$ by $scope(f)$.

DEFINITION 1 **(graphical model)** *(Kask, Dechter, Larrosa, & Dechter, 2005) A graphical model $\mathcal{M}$ is a 3-tuple, $\mathcal{M} = \langle \mathbf{X}, \mathbf{D}, \mathbf{F} \rangle$, where: $\mathbf{X} = \{X_1, \ldots, X_n\}$ is a finite set of variables; $\mathbf{D} = \{D_1, \ldots, D_n\}$ is the set of their respective finite domains of values; $\mathbf{F} = \{f_1, \ldots, f_r\}$ is a set of positive real-valued discrete functions, each defined over a subset of variables $\mathbf{S}_i \subseteq \mathbf{X}$, called its scope, and denoted by $scope(f_i)$. A graphical model typically has an associated combination operator [1] $\otimes$, (e.g., $\otimes \in \{\Pi, \sum\}$ - product, sum). The graphical model represents the combination*

---

1. The combination operator can also be defined axiomatically (Shenoy, 1992).

*of all its functions: $\otimes_{i=1}^{r} f_i$. A graphical model has an associated primal graph that captures the structural information of the model:*

**DEFINITION 2 (primal graph, dual graph)** *The* primal graph *of a graphical model is an undirected graph that has variables as its vertices and an edge connects any two vertices whose corresponding variables appear in the scope of the same function. A* dual graph *of a graphical model has a one-to-one mapping between its vertices and functions of the graphical model. Two vertices in the dual graph are connected if the corresponding functions in the graphical model share a variable. We denote the primal graph by $G = (X, E)$, where $X$ is the set of variables and $E$ is the set of edges.*

**DEFINITION 3 (belief networks)** *A* belief (or Bayesian) network *is a graphical model $\mathcal{B} = \langle X, D, G, P \rangle$, where $G = (X, E)$ is a directed acyclic graph over variables $X$ and $P = \{p_i\}$, where $p_i = \{p(X_i \mid pa(X_i))\}$ are conditional probability tables (CPTs) associated with each variable $X_i$ and $pa(X_i) = scope(p_i) - \{X_i\}$ is the set of parents of $X_i$ in G. Given a subset of variables $S$, we will write $P(s)$ as the probability $P(S = s)$, where $s \in D_S$. A belief network represents a probability distribution over $X$, $P(x_1, \ldots, x_n) = \Pi_{i=1}^{n} P(x_i | x_{pa(X_i)})$. An evidence set $e$ is an instantiated subset of variables. The primal graph of a belief network is called a moral graph. It can be obtained by connecting the parents of each vertex in G and removing the directionality of the edges. Equivalently, it connects any two variables appearing in the same family (a variable and its parents in the CPT).*

Two common queries in Bayesian networks are Belief Updating (BU) and Most Probable Explanation (MPE).

**DEFINITION 4 (belief network queries)** *The* Belief Updating (BU) *task is to find the posterior probability of each single variable given some evidence $e$, that is to compute $P(X_i|e)$. The* Most Probable Explanation (MPE) *task is to find a complete assignment to all the variables having maximum probability given the evidence, that is to compute $argmax_X \Pi_i p_i$.*

## 2.2 Tree-Decomposition Schemes

Tree-decomposition is at the heart of most general schemes for solving a wide range of automated reasoning problems, such as constraint satisfaction and probabilistic inference. It is the basis for many well-known algorithms, such as join-tree clustering and bucket elimination. In our presentation we will follow the terminology of Gottlob, Leone, and Scarcello (2000) and Kask et al. (2005).

**DEFINITION 5 (tree-decomposition, cluster-tree)** *Let $\mathcal{B} = \langle X, D, G, P \rangle$ be a belief network. A tree-decomposition for $\mathcal{B}$ is a triple $\langle T, \chi, \psi \rangle$, where $T = (V, E)$ is a tree, and $\chi$ and $\psi$ are labeling functions which associate with each vertex $v \in V$ two sets, $\chi(v) \subseteq X$ and $\psi(v) \subseteq P$ satisfying:*

1. *For each function $p_i \in P$, there is* exactly *one vertex $v \in V$ such that $p_i \in \psi(v)$, and $scope(p_i) \subseteq \chi(v)$.*
2. *For each variable $X_i \in X$, the set $\{v \in V | X_i \in \chi(v)\}$ induces a connected subtree of $T$. This is also called the running intersection (or connectedness) property.*

*We will often refer to a node and its functions as a* cluster *and use the term* tree-decomposition *and* cluster-tree *interchangeably.*

DEFINITION 6 (treewidth, separator, eliminator) *Let $D = \langle T, \chi, \psi \rangle$ be a tree-decomposition of a belief network $\mathcal{B}$. The* treewidth *(Arnborg, 1985) of $D$ is $max_{v \in V} |\chi(v)| - 1$. The treewidth of $\mathcal{B}$ is the minimum treewidth over all its tree-decompositions. Given two adjacent vertices $u$ and $v$ of a tree-decomposition, the* separator *of $u$ and $v$ is defined as $sep(u, v) = \chi(u) \cap \chi(v)$, and the* eliminator *of $u$ with respect to $v$ is $elim(u, v) = \chi(u) - \chi(v)$. The* separator-width *of $D$ is $max_{(u,v)} |sep(u, v)|$. The minimum treewidth of a graph $G$ can be shown to be identical to a related parameter called* induced-width *(Dechter & Pearl, 1987).*

**Join-tree and cluster-tree elimination (CTE)** In both Bayesian network and constraint satisfaction communities, the most used tree-decomposition method is join-tree decomposition (Lauritzen & Spiegelhalter, 1988; Dechter & Pearl, 1989), introduced based on relational database concepts (Maier, 1983). Such decompositions can be generated by embedding the network's moral graph $G$ into a chordal graph, often using a triangulation algorithm and using its maximal cliques as nodes in the join-tree. The triangulation algorithm assembles a join-tree by connecting the maximal cliques in the chordal graph in a tree. Subsequently, every CPT $p_i$ is placed in one clique containing its scope. Using the previous terminology, a join-tree decomposition of a belief network $\mathcal{B} = \langle X, D, G, P \rangle$ is a tree $T = (V, E)$, where $V$ is the set of cliques of a chordal graph $G'$ that contains $G$, and $E$ is a set of edges that form a tree between cliques, satisfying the running intersection property (Maier, 1983). Such a join-tree satisfies the properties of tree-decomposition and is therefore a cluster-tree (Kask et al., 2005). In this paper, we will use the terms tree-decomposition and join-tree decomposition interchangeably.

There are a few variants for processing join-trees for belief updating (e.g., Jensen et al., 1990; Shafer & Shenoy, 1990). We adopt here the version from Kask et al. (2005), called cluster-tree-elimination (CTE), that is applicable to tree-decompositions in general and is geared towards space savings. It is a message-passing algorithm; for the task of belief updating, messages are computed by summation over the eliminator between the two clusters of the product of functions in the originating cluster. The algorithm, denoted CTE-BU (see Figure 1), pays a special attention to the processing of observed variables since the presence of evidence is a central component in belief updating. When a cluster sends a message to a neighbor, the algorithm operates on all the functions in the cluster except the message from that particular neighbor. The message contains a single *combined* function and *individual* functions that do not share variables with the relevant eliminator. All the non-individual functions are *combined* in a product and summed over the eliminator.

**Example 1** *Figure 2a describes a belief network and Figure 2b a join-tree decomposition for it. Figure 2c shows the trace of running CTE-BU with evidence $G = g_e$, where $h_{(u,v)}$ is a message that cluster $u$ sends to cluster $v$.*

THEOREM 1 (complexity of CTE-BU) *(Dechter et al., 2001; Kask et al., 2005) Given a Bayesian network $\mathcal{B} = \langle X, D, G, P \rangle$ and a tree-decomposition $\langle T, \chi, \psi \rangle$ of $\mathcal{B}$, the time complexity of CTE-BU is $O(deg \cdot (n + N) \cdot d^{w^*+1})$ and the space complexity is $O(N \cdot d^{sep})$, where deg is the maximum degree of a node in the tree-decomposition, n is the number of variables, N is the number of nodes in the tree-decomposition, d is the maximum domain size of a variable, $w^*$ is the treewidth and sep is the maximum separator size.*

Algorithm **CTE for Belief-Updating (CTE-BU)**
**Input:** A tree-decomposition $\langle T, \chi, \psi \rangle$, $T = (V, E)$ for $\mathcal{B} = \langle X, D, G, P \rangle$. Evidence variables $var(e)$.
**Output:** An augmented tree whose nodes are clusters containing the original CPTs and the messages received from neighbors. $P(X_i, e)$, $\forall X_i \in X$.

Denote by $H_{(u,v)}$ the message from vertex $u$ to $v$, $ne_v(u)$ the neighbors of $u$ in $T$ excluding $v$,
$cluster(u) = \psi(u) \cup \{H_{(v,u)} | (v, u) \in E\}$,
$cluster_v(u) = cluster(u)$ excluding message from $v$ to $u$.

• **Compute messages:**
**For** every node $u$ in $T$, once $u$ has received messages from all $ne_v(u)$, compute message to node $v$:

    1. **Process observed variables:**
       Assign relevant evidence to all $p_i \in \psi(u)$
    2. **Compute the combined function:**

$$h_{(u,v)} = \sum_{elim(u,v)} \prod_{f \in A} f$$

    where $A$ is the set of functions in $cluster_v(u)$ whose scope intersects $elim(u,v)$.
    Add $h_{(u,v)}$ to $H_{(u,v)}$ and add all the individual functions in $cluster_v(u) - A$
    Send $H_{(u,v)}$ to node $v$.

• **Compute $P(X_i, e)$:**
For every $X_i \in X$ let $u$ be a vertex in $T$ such that $X_i \in \chi(u)$. Compute $P(X_i, e) = \sum_{\chi(u) - \{X_i\}} (\prod_{f \in cluster(u)} f)$

Figure 1: Algorithm Cluster-Tree-Elimination for Belief Updating (CTE-BU).

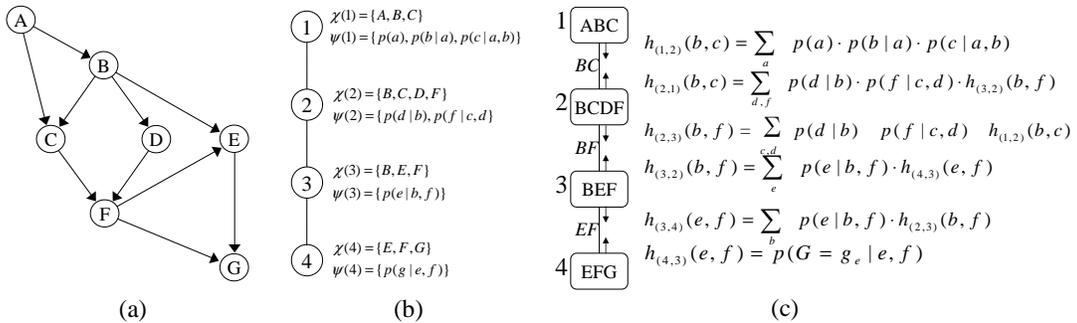

Figure 2: (a) A belief network; (b) A join-tree decomposition; (c) Execution of CTE-BU.

# 3. Partition-Based Mini-Clustering

The time, and especially the space complexity, of CTE-BU renders the algorithm infeasible for problems with large treewidth. We now introduce Mini-Clustering, a partition-based anytime algorithm which computes bounds or approximate values on $P(X_i, e)$ for every variable $X_i$.

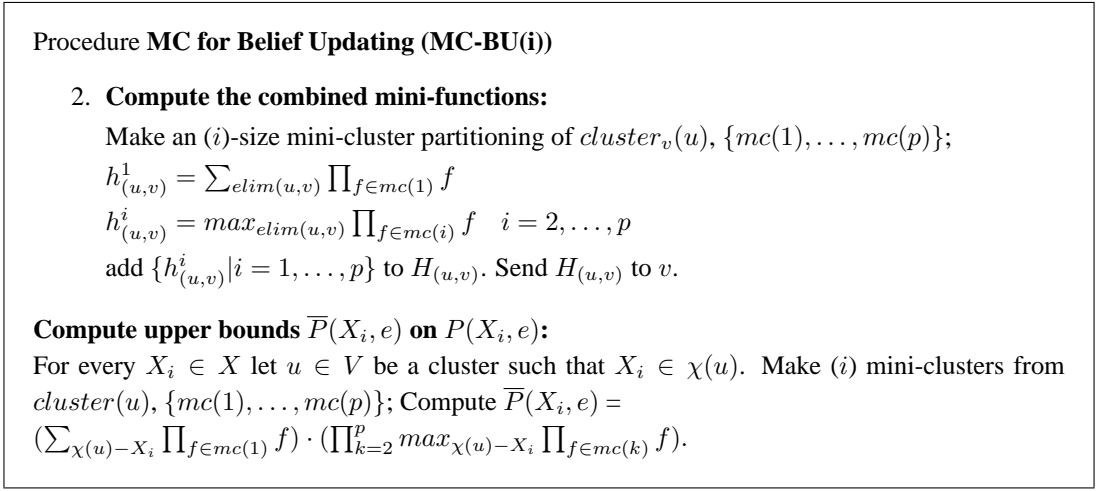

Procedure **MC for Belief Updating (MC-BU(i))**

2. **Compute the combined mini-functions:**

Make an ($i$)-size mini-cluster partitioning of $cluster_v(u)$, $\{mc(1), \ldots, mc(p)\}$;

$h^1_{(u,v)} = \sum_{elim(u,v)} \prod_{f \in mc(1)} f$

$h^i_{(u,v)} = max_{elim(u,v)} \prod_{f \in mc(i)} f \quad i = 2, \ldots, p$

add $\{h^i_{(u,v)} | i = 1, \ldots, p\}$ to $H_{(u,v)}$. Send $H_{(u,v)}$ to $v$.

**Compute upper bounds $\overline{P}(X_i, e)$ on $P(X_i, e)$:**

For every $X_i \in X$ let $u \in V$ be a cluster such that $X_i \in \chi(u)$. Make ($i$) mini-clusters from $cluster(u)$, $\{mc(1), \ldots, mc(p)\}$; Compute $\overline{P}(X_i, e) =$
$(\sum_{\chi(u) - X_i} \prod_{f \in mc(1)} f) \cdot (\prod^p_{k=2} max_{\chi(u) - X_i} \prod_{f \in mc(k)} f)$.

Figure 3: Procedure Mini-Clustering for Belief Updating (MC-BU).

## 3.1 Mini-Clustering Algorithm

Combining all the functions of a cluster into a product has a complexity exponential in its number of variables, which is upper bounded by the induced width. Similar to the mini-bucket scheme (Dechter, 1999), rather than performing this expensive exact computation, we partition the cluster into $p$ mini-clusters $mc(1), \ldots, mc(p)$, each having at most $i$ variables, where $i$ is an accuracy parameter. Instead of computing by CTE-BU $h_{(u,v)} = \sum_{elim(u,v)} \prod_{f \in \psi(u)} f$, we can divide the functions of $\psi(u)$ into $p$ mini-clusters $mc(k), k \in \{1, \ldots, p\}$, and rewrite $h_{(u,v)} = \sum_{elim(u,v)} \prod_{f \in \psi(u)} f = \sum_{elim(u,v)} \prod^p_{k=1} \prod_{f \in mc(k)} f$. By migrating the summation operator into each mini-cluster, yielding $\prod^p_{k=1} \sum_{elim(u,v)} \prod_{f \in mc(k)} f$, we get an upper bound on $h_{(u,v)}$. The resulting algorithm is called MC-BU(i).

Consequently, the combined functions are approximated via mini-clusters, as follows. Suppose $u \in V$ has received messages from all its neighbors other than $v$ (the message from $v$ is ignored even if received). The functions in $cluster_v(u)$ that are to be combined are partitioned into mini-clusters $\{mc(1), \ldots, mc(p)\}$, each one containing at most $i$ variables. Each mini-cluster is processed by summation over the eliminator, and the resulting combined functions as well as all the individual functions are sent to $v$. It was shown by Dechter and Rish (2003) that the upper bound can be improved by using the maximization operator *max* rather than the summation operator *sum* on some mini-buckets. Similarly, lower bounds can be generated by replacing *sum* with *min* (minimization) for some mini-buckets. Alternatively, we can replace *sum* by a *mean* operator (taking the sum and dividing by the number of elements in the sum), in this case deriving an approximation of the joint belief instead of a strict upper bound.

Algorithm MC-BU for upper bounds can be obtained from CTE-BU by replacing step 2 of the main loop and the final part of computing the upper bounds on the joint belief by the procedure given in Figure 3. In the implementation we used for the experiments reported here, the partitioning was done in a greedy brute-force manner. We ordered the functions according to their sizes (number of variables), breaking ties arbitrarily. The largest function was placed in a mini-cluster by itself. Then, we picked the largest remaining function and probed the mini-clusters in the order of their creation, trying to find one that together with the new function would have no more than $i$ variables. A new mini-cluster was created whenever the existing ones could not accommodate the new function.

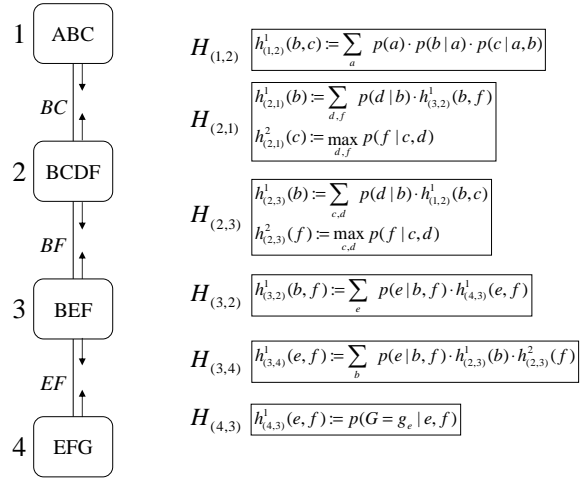

Figure 4: Execution of MC-BU for $i = 3$.

**Example 2** *Figure 4 shows the trace of running MC-BU(3) on the problem in Figure 2. First, evidence $G = g_e$ is assigned in all CPTs. There are no individual functions to be sent from cluster 1 to cluster 2. Cluster 1 contains only 3 variables, $\chi(1) = \{A, B, C\}$, therefore it is not partitioned. The combined function $h^1_{(1,2)}(b,c) = \sum_a p(a) \cdot p(b|a) \cdot p(c|a,b)$ is computed and the message $H_{(1,2)} = \{h^1_{(1,2)}(b,c)\}$ is sent to node 2. Now, node 2 can send its message to node 3. Again, there are no individual functions. Cluster 2 contains 4 variables, $\chi(2) = \{B, C, D, F\}$, and a partitioning is necessary: MC-BU(3) can choose $mc(1) = \{p(d|b), h_{(1,2)}(b,c)\}$ and $mc(2) = \{p(f|c,d)\}$. The combined functions $h^1_{(2,3)}(b) = \sum_{c,d} p(d|b) \cdot h_{(1,2)}(b,c)$ and $h^2_{(2,3)}(f) = max_{c,d} p(f|c,d)$ are computed and the message $H_{(2,3)} = \{h^1_{(2,3)}(b), h^2_{(2,3)}(f)\}$ is sent to node 3. The algorithm continues until every node has received messages from all its neighbors. An upper bound on $p(a, G = g_e)$ can now be computed by choosing cluster 1, which contains variable $A$. It doesn't need partitioning, so the algorithm just computes $\sum_{b,c} p(a) \cdot p(b|a) \cdot p(c|a,b) \cdot h^1_{(2,1)}(b) \cdot h^2_{(2,1)}(c)$. Notice that unlike CTE-BU which processes 4 variables in cluster 2, MC-BU(3) never processes more than 3 variables at a time.*

It was already shown that:

**THEOREM 2** *(Dechter & Rish, 2003) Given a Bayesian network $\mathcal{B} = \langle X, D, G, P \rangle$ and the evidence $e$, the algorithm MC-BU(i) computes an upper bound on the joint probability $P(X_i, e)$ of each variable $X_i$ (and each of its values) and the evidence $e$.*

**THEOREM 3 (complexity of MC-BU(i))** *(Dechter et al., 2001) Given a Bayesian network $\mathcal{B} = \langle X, D, G, P \rangle$ and a tree-decomposition $\langle T, \chi, \psi \rangle$ of $\mathcal{B}$, the time and space complexity of MC-BU(i) is $O(n \cdot hw^* \cdot d^i)$, where $n$ is the number of variables, $d$ is the maximum domain size of a variable and $hw^* = max_{u \in T} |\{f \in P | scope(f) \cap \chi(u) \neq \phi\}|$, which bounds the number of mini-clusters.*

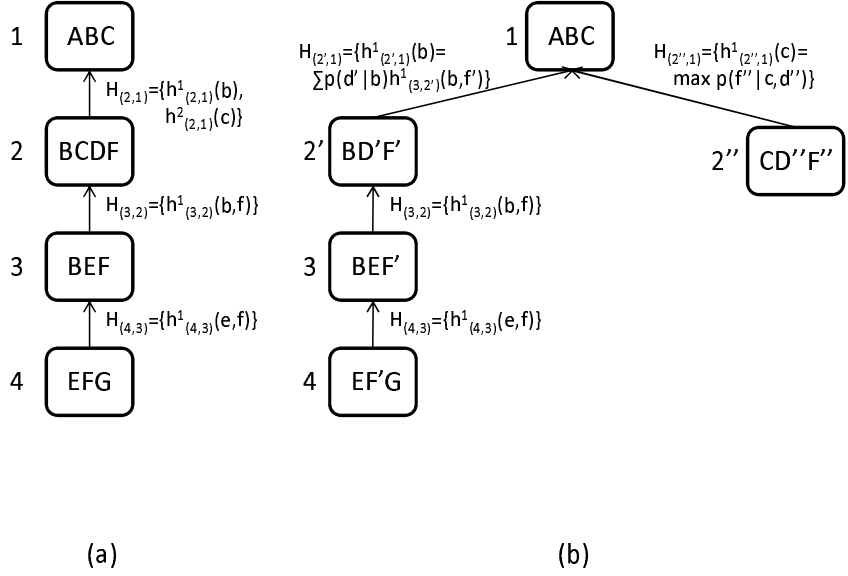

Figure 5: Node duplication semantics of MC: (a) trace of MC-BU(3); (b) trace of CTE-BU.

**Semantics of Mini-Clustering**  The mini-bucket scheme was shown to have the semantics of relaxation via *node duplication* (Kask & Dechter, 2001; Choi, Chavira, & Darwiche, 2007). We extend it to mini-clustering by showing how it can apply as is to messages that flow in one direction (inward, from leaves to root), as follows. Given a tree-decomposition $D$, where CTE-BU computes a function $h_{(u,v)}$ (the message that cluster $u$ sends to cluster $v$), MC-BU(i) partitions cluster $u$ into $p$ mini-clusters $u_1, \ldots, u_p$, which are processed independently and then the resulting functions $h_{(u_i,v)}$ are sent to $v$. Instead consider a different decomposition $D'$, which is just like $D$, with the exception that (a) instead of $u$, it has clusters $u_1, \ldots, u_p$, all of which are children of $v$, and each variable appearing in more than a single mini-cluster becomes a new variable, (b) each child $w$ of $u$ (in $D$) is a child of $u_k$ (in $D'$), such that $h_{(w,u)}$ (in $D$) is assigned to $u_k$ (in $D'$) during the partitioning. Note that $D'$ is not a legal tree-decomposition relative to the original variables since it violates the connectedness property: the mini-clusters $u_1, \ldots, u_p$ contain variables $elim(u, v)$ but the path between the nodes $u_1, \ldots, u_p$ (this path goes through $v$) does not. However, it is a legal tree-decomposition relative to the new variables. It is straightforward to see that $H_{(u,v)}$ computed by MC-BU(i) on $D$ is the same as $\{h_{(u_i,v)} | i = 1, \ldots, p\}$ computed by CTE-BU on $D'$ in the direction from leaves to root.

If we want to capture the semantics of the outward messages from root to leaves, we need to generate a different relaxed decomposition ($D''$) because MC, as defined, allows a different partitioning in the up and down streams of the same cluster. We could of course stick with the decomposition in $D'$ and use CTE in both directions which would lead to another variant of mini-clustering.

**Example 3**  *Figure 5(a) shows a trace of the bottom-up phase of MC-BU(3) on the network in Figure 4. Figure 5(b) shows a trace of the bottom-up phase of CTE-BU algorithm on a problem obtained from the problem in Figure 4 by splitting nodes $D$ (into $D'$ and $D''$) and $F$ (into $F'$ and $F''$).*

The MC-BU algorithm computes an upper bound $\overline{P}(X_i, e)$ on the joint probability $P(X_i, e)$. However, deriving a bound on the conditional probability $P(X_i|e)$ is not easy when the exact

RandomBayesianN=50K=2P=2C=48

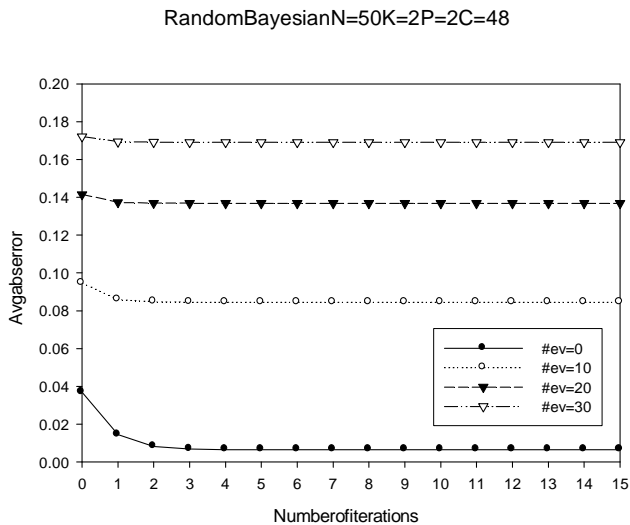

Figure 6: Convergence of IBP (50 variables, evidence from 0-30 variables).

value of $P(e)$ is not available. If we just try to divide (multiply) $\overline{P}(X_i, e)$ by a constant, the result is not necessarily an upper bound on $P(X_i|e)$. It is easy to show that normalization, $\overline{P}(x_i, e) / \sum_{x_i \in D_i} \overline{P}(x_i, e)$, with the $mean$ operator is identical to normalization of MC-BU output when applying the summation operator in all the mini-clusters.

MC-BU(i) is an improvement over the Mini-Bucket algorithm MB(i), in that it allows the computation of $\overline{P}(X_i, e)$ for all variables with a single run, whereas MB(i) computes $\overline{P}(X_i, e)$ for just one variable, with a single run. When computing $\overline{P}(X_i, e)$ for each variable, MB(i) has to be run $n$ times, once for each variable, an algorithm we call nMB(i). It was demonstrated by Mateescu et al. (2002) that MC-BU(i) has up to linear speed-up over nMB(i). For a given $i$, the accuracy of MC-BU(i) can be shown to be not worse than that of nMB(i).

### 3.2 Experimental Evaluation of Mini-Clustering

The work of Mateescu et al. (2002) and Kask (2001) provides an empirical evaluation of MC-BU that reveals the impact of the accuracy parameter on its quality of approximation and compares with Iterative Belief Propagation and a Gibbs sampling scheme. We will include here only a subset of these experiments which will provide the essence of our results. Additional empirical evaluation of MC-BU will be given when comparing against IJGP later in this paper.

We tested the performance of MC-BU(i) on random Noisy-OR networks, random coding networks, general random networks, grid networks, and three benchmark CPCS files with 54, 360 and 422 variables respectively (these are belief networks for medicine, derived from the Computer based Patient Case Simulation system, known to be hard for belief updating). On each type of network we ran Iterative Belief Propagation (IBP) - set to run at most 30 iterations, Gibbs Sampling (GS) and MC-BU(i), with $i$ from 2 to the treewidth $w^*$ to capture the anytime behavior of MC-BU(i).

The random networks were generated using parameters (N,K,C,P), where N is the number of variables, K is their domain size (we used only K=2), C is the number of conditional probability tables and P is the number of parents in each conditional probability table. The parents in each table are picked randomly given a topological ordering, and the conditional probability tables are filled randomly. The grid networks have the structure of a square, with edges directed to form a diagonal flow (all parallel edges have the same direction). They were generated by specifying N (a square integer) and K (we used K=2). We also varied the number of evidence nodes, denoted by $|e|$ in the tables. The parameter values are reported in each table. For all the problems, Gibbs sampling performed consistently poorly so we only include part of its results here.

| N=50, P=2, 50 instances | | | | | | | | |
|---|---|---|---|---|---|---|---|---|
| 0<br>\|e\| 10<br>20 | NHD | | Abs. Error | | Rel. Error | | Time | |
| | max | mean | max | mean | max | mean | max | mean |
| IBP | | 0 | | 9.0E-09 | | 1.1E-05 | | 0.102 |
| | | 0 | | 3.4E-04 | | 4.2E-01 | | 0.081 |
| | | 0 | | 9.6E-04 | | 1.2E+00 | | 0.062 |
| MC-BU(2) | 0 | 0 | 1.6E-03 | 1.1E-03 | 1.9E+00 | 1.3E+00 | 0.056 | 0.057 |
| | 0 | 0 | 1.1E-03 | 8.4E-04 | 1.4E+00 | 1.0E+00 | 0.048 | 0.049 |
| | 0 | 0 | 5.7E-04 | 4.8E-04 | 7.1E-01 | 5.9E-01 | 0.039 | 0.039 |
| MC-BU(5) | 0 | 0 | 1.1E-03 | 9.4E-04 | 1.4E+00 | 1.2E+00 | 0.070 | 0.072 |
| | 0 | 0 | 7.7E-04 | 6.9E-04 | 9.3E-01 | 8.4E-01 | 0.063 | 0.066 |
| | 0 | 0 | 2.8E-04 | 2.7E-04 | 3.5E-01 | 3.3E-01 | 0.058 | 0.057 |
| MC-BU(8) | 0 | 0 | 3.6E-04 | 3.2E-04 | 4.4E-01 | 3.9E-01 | 0.214 | 0.221 |
| | 0 | 0 | 1.7E-04 | 1.5E-04 | 2.0E-01 | 1.9E-01 | 0.184 | 0.190 |
| | 0 | 0 | 3.5E-05 | 3.5E-05 | 4.3E-02 | 4.3E-02 | 0.123 | 0.127 |

Table 1: Performance on Noisy-OR networks, $w^* = 10$: Normalized Hamming Distance, absolute error, relative error and time.

In our experiments we focused on the approximation power of MC-BU(i). We compared two versions of the algorithm. In the first version, for every cluster, we used the *max* operator in all its mini-clusters, except for one of them that was processed by summation. In the second version, we used the operator *mean* in all the mini-clusters. We investigated this second version of the algorithm for two reasons: (1) we compare MC-BU(i) with IBP and Gibbs sampling, both of which are also approximation algorithms, so it would not be possible to compare with a bounding scheme; (2) we observed in our experiments that, although the bounds improve as the i-bound increases, the quality of bounds computed by MC-BU(i) was still poor, with upper bounds being greater than 1 in many cases.[2] Notice that we need to maintain the *sum* operator for at least one of the mini-clusters. The *mean* operator simply performs summation and divides by the number of elements in the sum. For example, if $A, B, C$ are binary variables (taking 0 and 1), and $f(A, B, C)$ is the aggregated function of one mini-cluster, and $elim = \{A, B\}$, then computing the message $h(C)$ by the *mean* operator gives: $1/4 \sum_{A,B \in \{0,1\}} f(A, B, C)$.

We computed the exact solution and used three different measures of accuracy: 1) Normalized Hamming Distance (NHD) - we picked the most likely value for each variable for the approximate and for the exact, took the ratio between the number of disagreements and the total number of variables, and averaged over the number of problems that we ran for each class; 2) Absolute Error (Abs. Error) - is the absolute value of the difference between the approximate and the exact, averaged over all values (for each variable), all variables and all problems; 3) Relative Error (Rel. Error) - is the absolute value of the difference between the approximate and the exact, divided by the exact, averaged over all values (for each variable), all variables and all problems. For coding networks, we report only one measure, Bit Error Rate (BER). In terms of the measures defined above, BER is the normalized Hamming distance between the approximate (computed by an algorithm) and the actual input (which in the case of coding networks may be different from the solution given by exact algorithms), so we denote them differently to make this semantic distinction. We also report the time taken by each algorithm. For reported metrics (time, error, etc.) provided in the Tables, we give both mean and max values.

---

2. Wexler and Meek (2008) compared the upper/lower bounding properties of the mini-bucket on computing probability of evidence. Rollon and Dechter (2010) further investigated heuristic schemes for mini-bucket partitioning.

| N=50, P=3, 25 instances | | | | | | | |
|---|---|---|---|---|---|---|---|
| 10 $\|e\|$ 20 30 | NHD | | Abs. Error | | Rel. Error | | Time | |
| | max | mean | max | mean | max | mean | max | mean |
| IBP | | 0 | | 1.3E-04 | | 7.9E-01 | | 0.242 |
| | | 0 | | 3.6E-04 | | 2.2E+00 | | 0.184 |
| | | 0 | | 6.8E-04 | | 4.2E+00 | | 0.121 |
| MC-BU(2) | 0 | 0 | 1.3E-03 | 9.6E-04 | 8.2E+00 | 5.8E+00 | 0.107 | 0.108 |
| | 0 | 0 | 5.3E-04 | 4.0E-04 | 3.1E+00 | 2.4E+00 | 0.077 | 0.077 |
| | 0 | 0 | 2.3E-04 | 1.9E-04 | 1.4E+00 | 1.2E+00 | 0.064 | 0.064 |
| MC-BU(5) | 0 | 0 | 1.0E-03 | 8.3E-04 | 6.4E+00 | 5.1E+00 | 0.133 | 0.133 |
| | 0 | 0 | 4.6E-04 | 4.1E-04 | 2.7E+00 | 2.4E+00 | 0.104 | 0.105 |
| | 0 | 0 | 2.0E-04 | 1.9E-04 | 1.2E+00 | 1.2E+00 | 0.098 | 0.095 |
| MC-BU(8) | 0 | 0 | 6.6E-04 | 5.7E-04 | 4.0E+00 | 3.5E+00 | 0.498 | 0.509 |
| | 0 | 0 | 1.8E-04 | 1.8E-04 | 1.1E+00 | 1.0E+00 | 0.394 | 0.406 |
| | 0 | 0 | 3.4E-05 | 3.4E-05 | 2.1E-01 | 2.1E-01 | 0.300 | 0.308 |
| MC-BU(11) | 0 | 0 | 2.6E-04 | 2.4E-04 | 1.6E+00 | 1.5E+00 | 2.339 | 2.378 |
| | 0 | 0 | 3.8E-05 | 3.8E-05 | 2.3E-01 | 2.3E-01 | 1.421 | 1.439 |
| | 0 | 0 | 6.4E-07 | 6.4E-07 | 4.0E-03 | 4.0E-03 | 0.613 | 0.624 |
| MC-BU(14) | 0 | 0 | 4.2E-05 | 4.1E-05 | 2.5E-01 | 2.4E-01 | 7.805 | 7.875 |
| | 0 | 0 | 0 | 0 | 0 | 0 | 2.075 | 2.093 |
| | 0 | 0 | 0 | 0 | 0 | 0 | 0.630 | 0.638 |

Table 2: Performance on Noisy-OR networks, $w^* = 16$: Normalized Hamming Distance, absolute error, relative error and time.

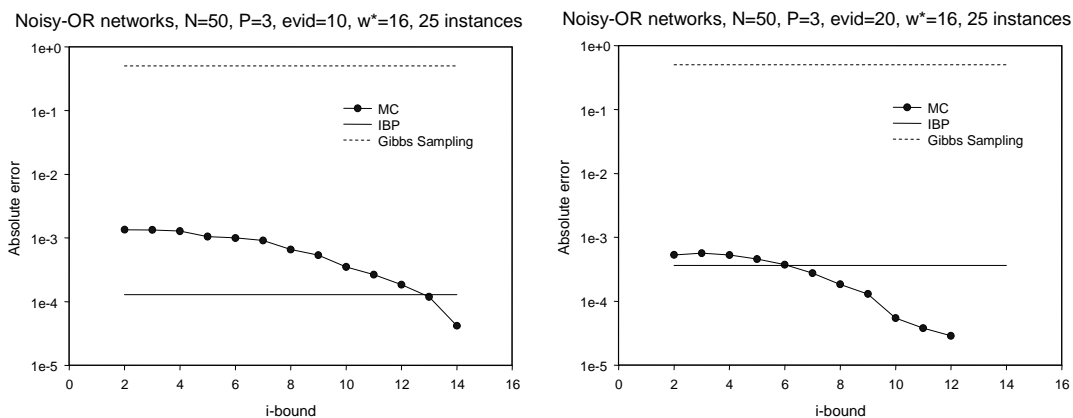

Figure 7: Absolute error for Noisy-OR networks.

In Figure 6 we show that IBP converges after about 5 iterations. So, while in our experiments we report its time for 30 iterations, its time is even better when sophisticated termination is used. These results are typical of all runs.

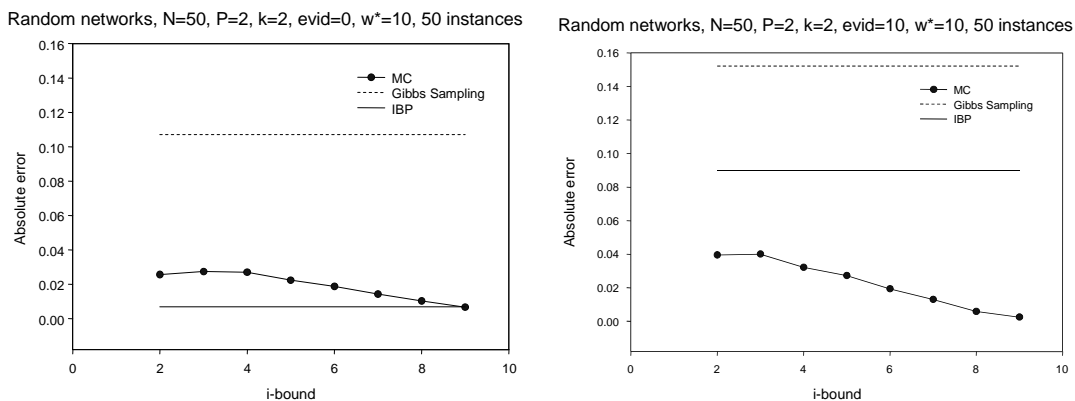

Figure 8: Absolute error for random networks.

| BER | $\sigma = .22$ | | $\sigma = .26$ | | $\sigma = .32$ | | $\sigma = .40$ | | $\sigma = .51$ | | Time |
|---|---|---|---|---|---|---|---|---|---|---|---|
| | max | mean | max | mean | max | mean | max | mean | max | mean | |
| N=100, P=3, 50 instances, w*=7 | | | | | | | | | | | |
| IBP | 0.000 | 0.000 | 0.000 | 0.000 | 0.002 | 0.002 | 0.022 | 0.022 | 0.088 | 0.088 | 0.00 |
| GS | 0.483 | 0.483 | 0.483 | 0.483 | 0.483 | 0.483 | 0.483 | 0.483 | 0.483 | 0.483 | 31.36 |
| MC-BU(2) | 0.002 | 0.002 | 0.004 | 0.004 | 0.024 | 0.024 | 0.068 | 0.068 | 0.132 | 0.131 | 0.08 |
| MC-BU(4) | 0.001 | 0.001 | 0.002 | 0.002 | 0.018 | 0.018 | 0.046 | 0.045 | 0.110 | 0.110 | 0.08 |
| MC-BU(6) | 0.000 | 0.000 | 0.000 | 0.000 | 0.004 | 0.004 | 0.038 | 0.038 | 0.106 | 0.106 | 0.12 |
| MC-BU(8) | 0.000 | 0.000 | 0.000 | 0.000 | 0.002 | 0.002 | 0.023 | 0.023 | 0.091 | 0.091 | 0.19 |
| N=100, P=4, 50 instances, w*=11 | | | | | | | | | | | |
| IBP | 0.000 | 0.000 | 0.000 | 0.000 | 0.002 | 0.002 | 0.013 | 0.013 | 0.075 | 0.075 | 0.00 |
| GS | 0.506 | 0.506 | 0.506 | 0.506 | 0.506 | 0.506 | 0.506 | 0.506 | 0.506 | 0.506 | 39.85 |
| MC-BU(2) | 0.006 | 0.006 | 0.015 | 0.015 | 0.043 | 0.043 | 0.093 | 0.094 | 0.157 | 0.157 | 0.19 |
| MC-BU(4) | 0.006 | 0.006 | 0.017 | 0.017 | 0.049 | 0.049 | 0.104 | 0.102 | 0.158 | 0.158 | 0.19 |
| MC-BU(6) | 0.005 | 0.005 | 0.011 | 0.011 | 0.035 | 0.034 | 0.071 | 0.074 | 0.151 | 0.150 | 0.29 |
| MC-BU(8) | 0.002 | 0.002 | 0.004 | 0.004 | 0.022 | 0.022 | 0.059 | 0.059 | 0.121 | 0.122 | 0.71 |
| MC-BU(10) | 0.001 | 0.001 | 0.001 | 0.001 | 0.008 | 0.008 | 0.033 | 0.032 | 0.101 | 0.102 | 1.87 |

Table 3: Bit Error Rate (BER) for coding networks.

**Random Noisy-OR networks results** are summarized in Tables 1 and 2, and Figure 7. For NHD, both IBP and MC-BU gave perfect results. For the other measures, we noticed that IBP is more accurate when there is no evidence by about an order of magnitude. However, as evidence is added, IBP's accuracy decreases, while MC-BU's increases and they give similar results. We see that MC-BU gets better as the accuracy parameter $i$ increases, which shows its anytime behavior.

**General random networks results** are summarized in Figure 8. They are similar to those for random Noisy-OR networks. Again, IBP has the best result only when the number of evidence variables is small. It is remarkable how quickly MC-BU surpasses the performance of IBP as evidence is added (for more, see the results of Mateescu et al., 2002).

**Random coding networks results** are given in Table 3 and Figure 9. The instances fall within the class of linear block codes, ($\sigma$ is the channel noise level). It is known that IBP is very accurate for this class. Indeed, these are the only problems that we experimented with where IBP outperformed MC-BU throughout. The anytime behavior of MC-BU can again be seen in the variation of numbers in each column and more vividly in Figure 9.

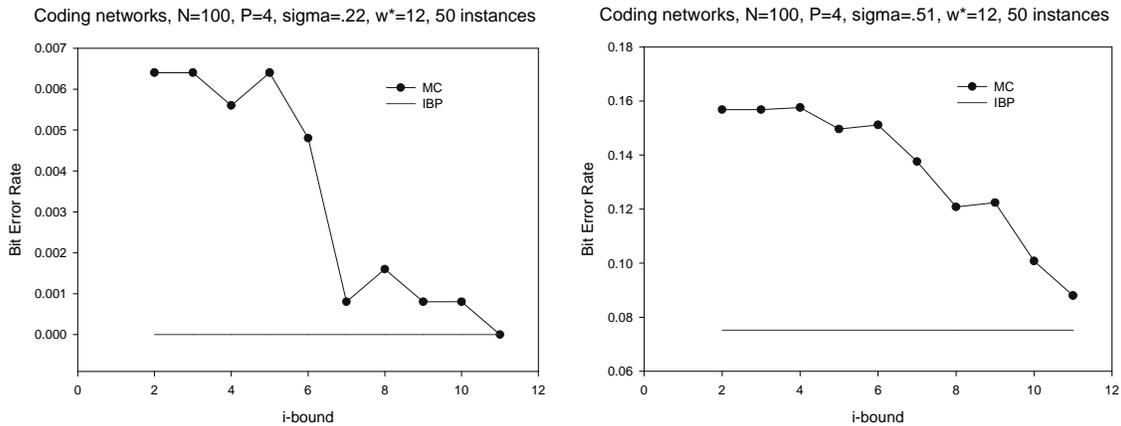

Figure 9: Bit Error Rate (BER) for coding networks.

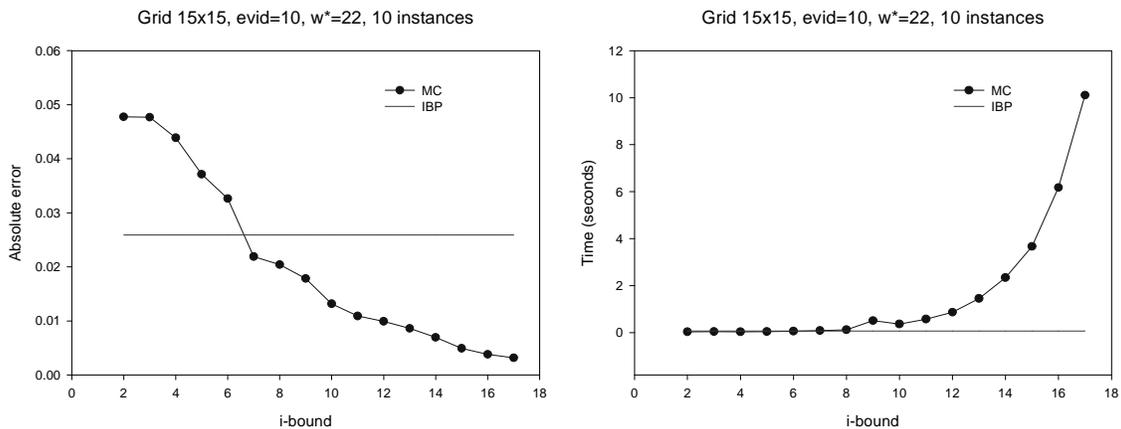

Figure 10: Grid 15x15: absolute error and time.

**Grid networks results**    are given in Figure 10. We notice that IBP is more accurate for no evidence and MC-BU is better as more evidence is added. The same behavior was consistently manifested for smaller grid networks that we experimented with (from 7x7 up to 14x14).

**CPCS networks results**    We also tested on three CPCS benchmark files. The results are given in Figure 11. It is interesting to notice that the MC-BU scheme scales up to fairly large networks, like the real life example of CPCS422 (induced width 23). IBP is again more accurate when there is no evidence, but is surpassed by MC-BU when evidence is added. However, whereas MC-BU is competitive with IBP time-wise when i-bound is small, its runtime grows rapidly as i-bound increases. For more details on all these benchmarks see the results of Mateescu et al. (2002).

**Summary**    Our results show that, as expected, IBP is superior to all other approximations for coding networks. However, for random Noisy-OR, general random, grid networks and the CPCS networks, in the presence of evidence, the mini-clustering scheme is often superior even in its weakest form. The empirical results are particularly encouraging as we use an un-optimized scheme that exploits a universal principle applicable to many reasoning tasks.

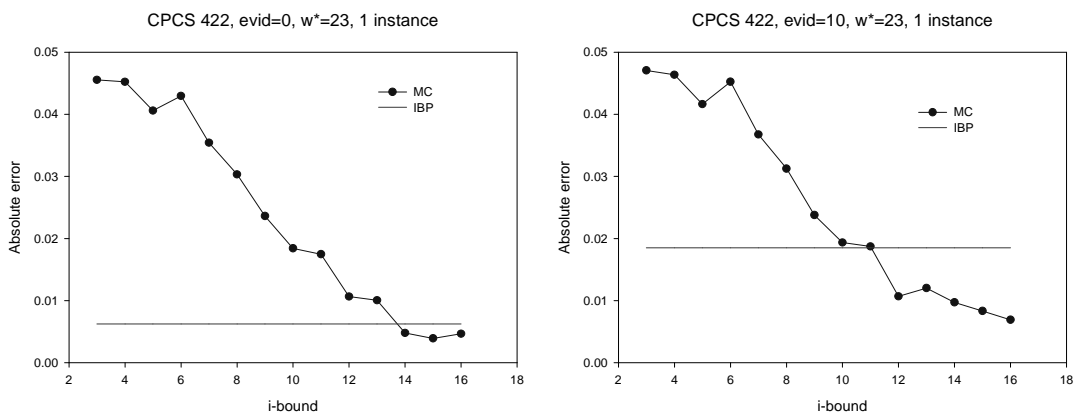

Figure 11: Absolute error for CPCS422.

# 4. Join-Graph Decomposition and Propagation

In this section we introduce algorithm Iterative Join-Graph Propagation (IJGP) which, like mini-clustering, is designed to benefit from bounded inference, but also exploit iterative message-passing as used by IBP. Algorithm IJGP can be viewed as an iterative version of mini-clustering, improving the quality of approximation, especially for low $i$-bounds. Given a cluster of the decomposition, mini-clustering can potentially create a different partitioning for every message sent to a neighbor. This dynamic partitioning can happen because the incoming message from each neighbor has to be excluded when realizing the partitioning, so a different set of functions are split into mini-clusters for every message to a neighbor. We can define a version of mini-clustering where for every cluster we create a unique static partition into mini-clusters such that every incoming message can be included into one of the mini-clusters. This version of MC can be extended into IJGP by introducing some links between mini-clusters of the same cluster, and carefully limiting the interaction between the resulting nodes in order to eliminate over-counting.

Algorithm IJGP works on a general join-graph that may contain cycles. The cluster size of the graph is user adjustable via the *i-bound* (providing the anytime nature), and the cycles in the graph allow the iterative application of message-passing. In Subsection 4.1 we introduce join-graphs and discuss their properties. In Subsection 4.2 we describe the IJGP algorithm itself.

## 4.1 Join-Graphs

DEFINITION 7 (**join-graph decomposition**) *A join-graph decomposition for a belief network* $\mathcal{B} = \langle X, D, G, P \rangle$ *is a triple* $D = \langle JG, \chi, \psi \rangle$, *where* $JG = (V, E)$ *is a graph, and* $\chi$ *and* $\psi$ *are labeling functions which associate with each vertex* $v \in V$ *two sets,* $\chi(v) \subseteq X$ *and* $\psi(v) \subseteq P$ *such that:*

1. *For each* $p_i \in P$, *there is* exactly *one vertex* $v \in V$ *such that* $p_i \in \psi(v)$, *and* $scope(p_i) \subseteq \chi(v)$.
2. *(connectedness) For each variable* $X_i \in X$, *the set* $\{v \in V | X_i \in \chi(v)\}$ *induces a connected subgraph of* $JG$. *The connectedness requirement is also called the running intersection property.*

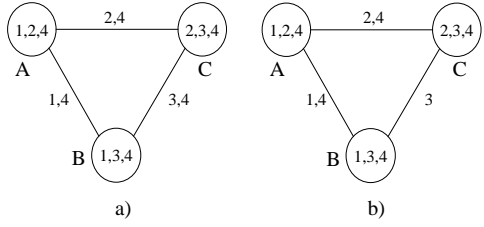

Figure 12: An edge-labeled decomposition.

We will often refer to a node in $V$ and its CPT functions as a *cluster*[3] and use the term *join-graph decomposition* and *cluster-graph* interchangeably. Clearly, a *join-tree decomposition* or a *cluster-tree* is the special case when the join-graph $D$ is a tree.

It is clear that one of the problems of message propagation over cyclic join-graphs is *over-counting*. To reduce this problem we devise a scheme, which avoids cyclicity with respect to any single variable. The algorithm works on edge-labeled join-graphs.

**DEFINITION 8 (minimal edge-labeled join-graph decompositions)** *An edge-labeled join-graph decomposition for* $\mathcal{B} = \langle X, D, G, P \rangle$ *is a four-tuple* $D = \langle JG, \chi, \psi, \theta \rangle$, *where* $JG = (V, E)$ *is a graph, $\chi$ and $\psi$ associate with each vertex $v \in V$ the sets $\chi(v) \subseteq X$ and $\psi(v) \subseteq P$ and $\theta$ associates with each edge $(v, u) \subset E$ the set $\theta((v, u)) \subseteq X$ such that:*

1. *For each function $p_i \in P$, there is exactly one vertex $v \in V$ such that $p_i \in \psi(v)$, and $scope(p_i) \subseteq \chi(v)$.*
2. *(edge-connectedness) For each edge $(u, v)$, $\theta((u, v)) \subseteq \chi(u) \cap \chi(v)$, such that $\forall X_i \in X$, any two clusters containing $X_i$ can be connected by a path whose every edge label includes $X_i$.*

*Finally, an edge-labeled join-graph is* minimal *if no variable can be deleted from any label while still satisfying the edge-connectedness property.*

**DEFINITION 9 (separator, eliminator of edge-labeled join-graphs)** *Given two adjacent vertices $u$ and $v$ of $JG$, the* separator *of $u$ and $v$ is defined as $sep(u, v) = \theta((u, v))$, and the* eliminator *of $u$ with respect to $v$ is $elim(u, v) = \chi(u) - \theta((u, v))$. The separator width is $\max_{(u,v)} |sep(u, v)|$.*

Edge-labeled join-graphs can be made label minimal by deleting variables from the labels while maintaining connectedness (if an edge label becomes empty, the edge can be deleted). It is easy to see that,

**Proposition 1** *A minimal edge-labeled join-graph* does not contain any cycle relative to any single variable. That is, any two clusters containing the same variable are connected by exactly one path labeled with that variable.

Notice that every minimal edge-labeled join-graph is edge-minimal (no edge can be deleted), but not vice-versa.

---

**Example 4** *The example in Figure 12a shows an edge minimal join-graph which contains a cycle relative to variable 4, with edges labeled with separators. Notice however that if we remove variable 4 from the label of one edge we will have no cycles (relative to single variables) while the connectedness property is still maintained.*

The mini-clustering approximation presented in the previous section works by relaxing the join-tree requirement of exact inference into a collection of join-trees having smaller cluster size. It introduces some independencies in the original problem via node duplication and applies exact inference on the relaxed model requiring only two message passings. For the class of IJGP algorithms we take a different route. We choose to relax the tree-structure requirement and use join-graphs which do not introduce any new independencies, and apply iterative message-passing on the resulting cyclic structure.

Indeed, it can be shown that any join-graph of a belief network is an I-map (independency map, Pearl, 1988) of the underlying probability distribution relative to node-separation. Since we plan to use minimally edge-labeled join-graphs to address over-counting problems, the question is what kind of independencies are captured by such graphs.

DEFINITION **10 (edge-separation in edge-labeled join-graphs)** *Let $D = \langle JG, \chi, \psi, \theta \rangle$, $JG = (V, E)$ be an edge-labeled decomposition of a Bayesian network $\mathcal{B} = \langle X, D, G, P \rangle$. Let $N_W, N_Y \subseteq V$ be two sets of nodes, and $E_Z \subseteq E$ be a set of edges in $JG$. Let $W, Y, Z$ be their corresponding sets of variables ($W = \cup_{v \in N_W} \chi(v)$, $Z = \cup_{e \in E_Z} \theta(e)$). We say that $E_Z$ edge-separates $N_W$ and $N_Y$ in $D$ if there is no path between $N_W$ and $N_Y$ in the $JG$ graph whose edges in $E_Z$ are removed. In this case we also say that $W$ is* separated *from $Y$ given $Z$ in $D$, and write $\langle W|Z|Y \rangle_D$. Edge-separation in a regular join-graph is defined relative to its separators.*

THEOREM **4** *Any edge-labeled join-graph decomposition $D = \langle JG, \chi, \psi, \theta \rangle$ of a belief network $\mathcal{B} = \langle X, D, G, P \rangle$ is an I-map of $P$ relative to edge-separation. Namely, any edge separation in $D$ corresponds to conditional independence in $P$.*

**Proof:** Let $MG$ be the moral graph of $BN$. Since $MG$ is an I-map of $P$, it is enough to prove that $JG$ is an I-map of $MG$. Let $N_W$ and $N_Y$ be disjoint sets of nodes and $N_Z$ be a set of edges in $JG$, and $W, Z, Y$ be their corresponding sets of variables in $MG$. We will prove:

$$\langle N_W|N_Z|N_Y \rangle_{JG} \Longrightarrow \langle W|Z|Y \rangle_{MG}$$

by contradiction. Since the sets $W, Z, Y$ may not be disjoint, we will actually prove that $\langle W - Z|Z|Y - Z \rangle_{MG}$ holds, this being equivalent to $\langle W|Z|Y \rangle_{MG}$.

Supposing $\langle W - Z|Z|Y - Z \rangle_{MG}$ is false, then there exists a path $\alpha = \gamma_1, \gamma_2, \ldots, \gamma_{n-1}, \beta = \gamma_n$ in $MG$ that goes from some variable $\alpha = \gamma_1 \in W - Z$ to some variable $\beta = \gamma_n \in Y - Z$ without intersecting variables in $Z$. Let $N_v$ be the set of all nodes in $JG$ that contain variable $v \in X$, and let us consider the set of nodes:

$$S = \cup_{i=1}^n N_{\gamma_i} - N_Z$$

We argue that $S$ forms a connected sub-graph in $JG$. First, the running intersection property ensures that every $N_{\gamma_i}, i = 1, \ldots, n$, remains connected in $JG$ after removing the nodes in $N_Z$ (otherwise, it must be that there was a path between the two disconnected parts in the original $JG$, which implies that a $\gamma_i$ is part of $Z$, which is a contradiction). Second, the fact that $(\gamma_i, \gamma_{i+1}), i =$ $1, \ldots, n-1$, is an edge in the moral graph $MG$ implies that there is a conditional probability table (CPT) on both $\gamma_i$ and $\gamma_{i+1}, i = 1, \ldots, n-1$ (and perhaps other variables). From property 1 of the definition of the join-graph, it follows that for all $i = 1, \ldots, n-1$ there exists a node in JG that contains both $\gamma_i$ and $\gamma_{i+1}$. This proves the existence of a path in the mutilated join-graph (JG with $N_Z$ pulled out) from a node in $N_W$ containing $\alpha = \gamma_1$ to the node containing both $\gamma_1$ and $\gamma_2$ ($N_{\gamma_1}$ is connected), then from that node to the one containing both $\gamma_2$ and $\gamma_3$ ($N_{\gamma_2}$ is connected), and so on until we reach a node in $N_Y$ containing $\beta = \gamma_n$. This shows that $\langle N_W | N_Z | N_Y \rangle_{JG}$ is false, concluding the proof by contradiction. $\qquad \square$

Interestingly however, deleting variables from edge labels or removing edges from edge-labeled join-graphs whose clusters are fixed will not increase the independencies captured by edge-labeled join-graphs. That is,

**Proposition 2** *Any two (edge-labeled) join-graphs defined on the same set of clusters, sharing ($V$, $\chi$, $\psi$), express exactly the same set of independencies relative to edge-separation, and this set of independencies is identical to the one expressed by node separation in the primal graph of the join-graph.*

**Proof:** This follows by looking at the primal graph of the join-graph (obtained by connecting any two nodes in a cluster by an arc over the original variables as nodes) and observing that any edge-separation in a join-graph corresponds to a node separation in the primal graph and vice-versa. $\qquad \square$

Hence, the issue of minimizing computational over-counting due to cycles appears to be unrelated to the problem of maximizing independencies via minimal I-mapness. Nevertheless, to avoid over-counting as much as possible, we still prefer join-graphs that minimize cycles relative to each variable. That is, we prefer *minimal* edge-labeled join-graphs.

**Relationship with region graphs**   There is a strong relationship between our join-graphs and the region graphs of Yedidia et al. (2000, 2001, 2005). Their approach was inspired by advances in statistical physics, when it was realized that computing the partition function is essentially the same combinatorial problem that expresses probabilistic reasoning. As a result, variational methods from physics could have counterparts in reasoning algorithms. It was proved by Yedidia et al. (2000, 2001) that belief propagation on loopy networks can only converge (when it does so) to stationary points of the Bethe free energy. The Bethe approximation is only the simplest case of the more general Kikuchi (1951) cluster variational method. The idea is to group the variables together in clusters and perform exact computation in each cluster. One key question is then how to aggregate the results, and how to account for the variables that are shared between clusters. Again, the idea that everything should be counted exactly once is very important. This led to the proposal of region graphs (Yedidia et al., 2001, 2005) and the associated counting numbers for regions. They are given as a possible canonical version of graphs that can support Generalized Belief Propagation (GBP) algorithms. The join-graphs accomplish the same thing. The edge-labeled join-graphs can be described as region graphs where the regions are the clusters and the labels on the edges. The tree-ness condition with respect to every variable ensures that there is no over-counting.

A very similar approach to ours, which is also based on join-graphs appeared independently by McEliece and Yildirim (2002), and it is based on an information theoretic perspective.

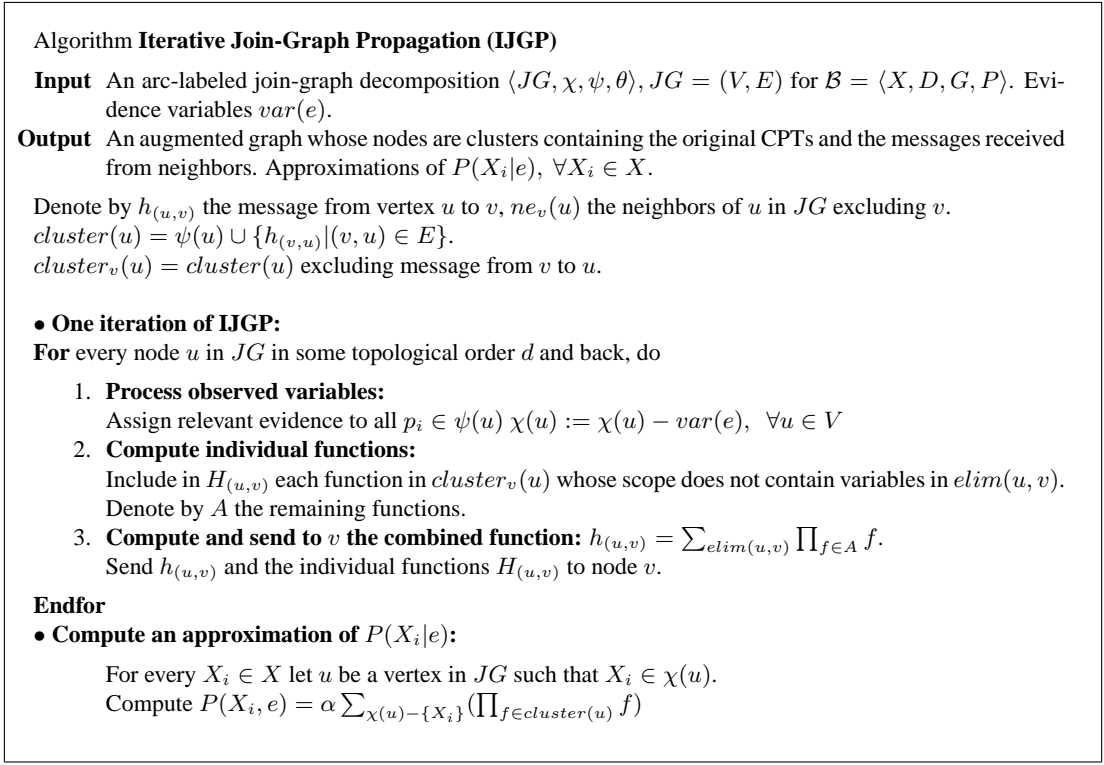

Figure 13: Algorithm Iterative Join-Graph Propagation (IJGP).

## 4.2 Algorithm IJGP

Applying CTE iteratively to minimal edge-labeled join-graphs yields our algorithm *Iterative Join-Graph Propagation (IJGP)* described in Figure 13. One iteration of the algorithm applies message-passing in a topological order over the join-graph, forward and back. When node $u$ sends a message (or messages) to a neighbor node $v$ it operates on all the CPTs in its cluster and on all the messages sent from its neighbors excluding the ones received from $v$. First, all individual functions that share no variables with the eliminator are collected and sent to $v$. All the rest of the functions are *combined* in a product and summed over the eliminator between $u$ and $v$.

Based on the results by Lauritzen and Spiegelhalter (1988) and Larrosa, Kask, and Dechter (2001) it can be shown that:

**THEOREM 5**    *1. If $IJGP$ is applied to a join-tree decomposition it reduces to join-tree clustering, and therefore it is guaranteed to compute the exact beliefs in one iteration.*

*2. The time complexity of one iteration of IJGP is $O(deg \cdot (n + N) \cdot d^{w^*+1})$ and its space complexity is $O(N \cdot d^\theta)$, where deg is the maximum degree of a node in the join-graph, n is the number of variables, N is the number of nodes in the graph decomposition, d is the maximum domain size, $w^*$ is the maximum cluster size and $\theta$ is the maximum label size.*

For proof, see the properties of CTE presented by Kask et al. (2005).

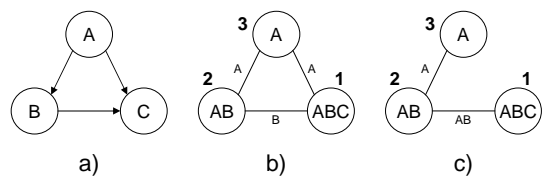

Figure 14: a) A belief network; b) A dual join-graph with singleton labels; c) A dual join-graph which is a join-tree.

**The special case of Iterative Belief Propagation**  Iterative belief propagation (IBP) is an iterative application of Pearl's algorithm that was defined for poly-trees (Pearl, 1988), to any Bayesian network. We will describe IBP as an instance of join-graph propagation over a *dual join-graph*.

DEFINITION **11 (dual graphs, dual join-graphs)** *Given a set of functions $F = \{f_1, \ldots, f_l\}$ over scopes $S_1, \ldots, S_l$, the dual graph of $F$ is a graph $DG = (V, E, L)$ that associates a node with each function, namely $V = F$ and an edge connects any two nodes whose function's scope share a variable, $E = \{(f_i, f_j) | S_i \cap S_j \neq \phi\}$. $L$ is a set of labels for the edges, each edge being labeled by the shared variables of its nodes, $L = \{l_{ij} = S_i \cap S_j | (i, j) \in E\}$. A dual join-graph is an edge-labeled edge subgraph of $DG$ that satisfies the connectedness property. A minimal dual join-graph is a dual join-graph for which none of the edge labels can be further reduced while maintaining the connectedness property.*

Interestingly, there may be many minimal dual join-graphs of the same dual graph. We will define Iterative Belief Propagation on any dual join-graph. Each node sends a message over an edge whose scope is identical to the label on that edge. Since Pearl's algorithm sends messages whose scopes are singleton variables only, we highlight minimal singleton-label dual join-graphs.

**Proposition 3** *Any Bayesian network has a minimal dual join-graph where each edge is labeled by a single variable.*

**Proof:** Consider a topological ordering of the nodes in the acyclic directed graph of the Bayesian network $d = X_1, \ldots, X_n$. We define the following dual join-graph. Every node in the dual graph $\mathcal{D}$, associated with $p_i$ is connected to node $p_j$, $j < i$ if $X_j \in pa(X_i)$. We label the edge between $p_j$ and $p_i$ by variable $X_j$, namely $l_{ij} = \{X_j\}$. It is easy to see that the resulting edge-labeled subgraph of the dual graph satisfies connectedness. (Take the original acyclic graph $G$ and add to each node its CPT family, namely all the other parents that precede it in the ordering. Since $G$ already satisfies connectedness so is the minimal graph generated.) The resulting labeled graph is a dual graph with singleton labels.  □

**Example 5** *Consider the belief network on 3 variables $A, B, C$ with CPTs 1.$P(C|A, B)$, 2.$P(B|A)$ and 3.$P(A)$, given in Figure 14a. Figure 14b shows a dual graph with singleton labels on the edges. Figure 14c shows a dual graph which is a join-tree, on which belief propagation can solve the problem exactly in one iteration (two passes up and down the tree).*

---

Algorithm **Iterative Belief Propagation (IBP)**

**Input:** An edge-labeled dual join-graph $DG = (V, E, L)$ for a Bayesian network $\mathcal{B} = \langle X, D, G, P \rangle$. Evidence $e$.

**Output:** An augmented graph whose nodes include the original CPTs and the messages received from neighbors. Approximations of $P(X_i|e)$, $\forall X_i \in X$. Approximations of $P(F_i|e)$, $\forall F_i \in \mathcal{B}$.

Denote by: $h_u^v$ the message from $u$ to $v$; $ne(u)$ the neighbors of $u$ in $V$; $ne_v(u) = ne(u) - \{v\}$; $l_{uv}$ the label of $(u, v) \in E$; $elim(u, v) = scope(u) - scope(v)$.

- **One iteration of IBP**
  **For** every node $u$ in $DJ$ in a topological order and back, do:
  1. **Process observed variables**
     Assign evidence variables to the each $p_i$ and remove them from the labeled edges.
  2. **Compute and send to $v$ the function:**

$$h_u^v = \sum_{elim(u,v)} (p_u \cdot \prod_{\{h_i^u, i \in ne_v(u)\}} h_i^u)$$

  **Endfor**
- **Compute approximations of** $P(F_i|e)$**,** $P(X_i|e)$**:**
  For every $X_i \in X$ let $u$ be the vertex of family $F_i$ in $DJ$,
  $P(F_i, e) = \alpha(\prod_{h_i^u, u \in ne(i)} h_i^u) \cdot p_u$;
  $P(X_i, e) = \alpha \sum_{scope(u) - \{X_i\}} P(F_i, e)$.

---

Figure 15: Algorithm Iterative Belief Propagation (IBP).

For completeness, we present algorithm IBP, which is a special case of IJGP, in Figure 15. It is easy to see that one iteration of IBP is time and space linear in the size of the belief network. It can be shown that when IBP is applied to a minimal singleton-labeled dual graph it coincides with Pearl's belief propagation applied directly to the acyclic graph representation. Also, when the dual join-graph is a tree IBP converges after one iteration (two passes, up and down the tree) to the exact beliefs.

### 4.3 Bounded Join-Graph Decompositions

Since we want to control the complexity of join-graph algorithms, we will define it on decompositions having bounded cluster size. If the number of variables in a cluster is bounded by $i$, the time and space complexity of processing one cluster is exponential in $i$. Given a join-graph decomposition $D = \langle JG, \chi, \psi, \theta \rangle$, the accuracy and complexity of the (iterative) join-graph propagation algorithm depends on two different width parameters, defined next.

DEFINITION **12 (external and internal widths)** *Given an edge-labeled join-graph decomposition* $D = \langle JG, \chi, \psi, \theta \rangle$ *of a network* $\mathcal{B} = \langle X, D, G, P \rangle$, *the* internal width *of $D$ is* $max_{v \in V}|\chi(v)|$, *while the* external width *of $D$ is the treewidth of $JG$ as a graph.*

Using this terminology we can now state our target decomposition more clearly. Given a graph $G$, and a bounding parameter $i$ we wish to find a join-graph decomposition $D$ of $G$ whose internal width is bounded by $i$ and whose external width is minimized. The bound $i$ controls the complexity of join-graph processing while the external width provides some measure of its accuracy and speed of convergence, because it measures how close the join-graph is to a join-tree.

---

Algorithm **Join-Graph Structuring($i$)**

1. Apply procedure schematic mini-bucket($i$).

2. Associate each resulting mini-bucket with a node in the join-graph, the variables of the nodes are those appearing in the mini-bucket, the original functions are those in the mini-bucket.

3. Keep the edges created by the procedure (called out-edges) and label them by the regular separator.

4. Connect the mini-bucket clusters belonging to the same bucket in a chain by in-edges labeled by the single variable of the bucket.

---

Figure 16: Algorithm Join-Graph Structuring($i$).

---

Procedure **Schematic Mini-Bucket($i$)**

1. Order the variables from $X_1$ to $X_n$ minimizing (heuristically) induced-width, and associate a bucket for each variable.

2. Place each CPT in the bucket of the highest index variable in its scope.

3. For $j = n$ to 1 do:
   Partition the functions in $bucket(X_j)$ into mini-buckets having at most $i$ variables.
   For each mini-bucket $mb$ create a new scope-function (message) $f$ where $scope(f) = \{X | X \in mb\} - \{X_i\}$ and place scope(f) in the bucket of its highest variable. Maintain an edge between $mb$ and the mini-bucket (created later) of $f$.

---

Figure 17: Procedure Schematic Mini-Bucket($i$).

We can consider two classes of algorithms. One class is *partition-based*. It starts from a given tree-decomposition and then partitions the clusters until the decomposition has clusters bounded by $i$. An alternative approach is *grouping-based*. It starts from a minimal dual-graph-based join-graph decomposition (where each cluster contains a single CPT) and groups clusters into larger clusters as long as the resulting clusters do not exceed the given bound. In both methods one should attempt to reduce the external width of the generated graph-decomposition. Our partition-based approach inspired by the mini-bucket idea (Dechter & Rish, 1997) is as follows.

Given a bound $i$, algorithm *Join-Graph Structuring(i)* applies the procedure *Schematic Mini-Bucket(i)*, described in Figure 17. The procedure only traces the scopes of the functions that would be generated by the full mini-bucket procedure, avoiding actual computation. The procedure ends with a collection of mini-bucket trees, each rooted in the mini-bucket of the first variable. Each of these trees is minimally edge-labeled. Then, *in-edges* labeled with only one variable are introduced, and they are added only to obtain the running intersection property between branches of these trees.

**Proposition 4** *Algorithm Join-Graph Structuring(i) generates a minimal edge-labeled join-graph decomposition having bound $i$.*

**Proof:** The construction of the join-graph specifies the vertices and edges of the join-graph, as well as the variable and function labels of each vertex. We need to demonstrate that 1) the connectedness property holds, and 2) that edge-labels are minimal.

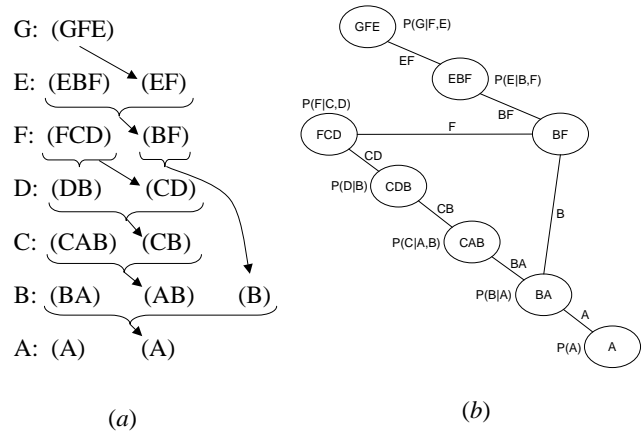

Figure 18: Join-graph decompositions.

Connectedness property specifies that for any 2 vertices $u$ and $v$, if vertices $u$ and $v$ contain variable $X$, then there must be a path $u, w_1, \ldots, w_m, v$ between $u$ and $v$ such that every vertex on this path contains variable $X$. There are two cases here. 1) $u$ and $v$ correspond to 2 mini-buckets in the same bucket, or 2) $u$ and $v$ correspond to mini-buckets in different buckets. In case 1 we have 2 further cases, 1a) variable $X$ is being eliminated in this bucket, or 1b) variable $X$ is not eliminated in this bucket. In case 1a, each mini-bucket must contain $X$ and all mini-buckets of the bucket are connected as a chain, so the connectedness property holds. In case 1b, vertexes $u$ and $v$ connect to their (respectively) parents, who in turn connect to their parents, etc. until a bucket in the scheme where variable $X$ is eliminated. All nodes along this chain connect variable $X$, so the connectedness property holds. Case 2 resolves like case 1b.

To show that edge labels are minimal, we need to prove that there are no cycles with respect to edge labels. If there is a cycle with respect to variable $X$, then it must involve at least one in-edge (edge connecting two mini-buckets in the same bucket). This means variable $X$ must be the variable being eliminated in the bucket of this in-edge. That means variable $X$ is not contained in any of the parents of the mini-buckets of this bucket. Therefore, in order for the cycle to exist, another in-edge down the bucket-tree from this bucket must contain $X$. However, this is impossible as this would imply that variable $X$ is eliminated twice. □

**Example 6** *Figure 18a shows the trace of procedure schematic mini-bucket(3) applied to the problem described in Figure 2a. The decomposition in Figure 18b is created by the algorithm graph structuring. The only cluster partitioned is that of F into two scopes (FCD) and (BF), connected by an in-edge labeled with F.*

A range of edge-labeled join-graphs is shown in Figure 19. On the left side we have a graph with smaller clusters, but more cycles. This is the type of graph IBP works on. On the right side we have a tree decomposition, which has no cycles at the expense of bigger clusters. In between, there could be a number of join-graphs where maximum cluster size can be traded for number of cycles. Intuitively, the graphs on the left present less complexity for join-graph algorithms because the cluster size is smaller, but they are also likely to be less accurate. The graphs on the right side are computationally more complex, because of the larger cluster size, but they are likely to be more accurate.

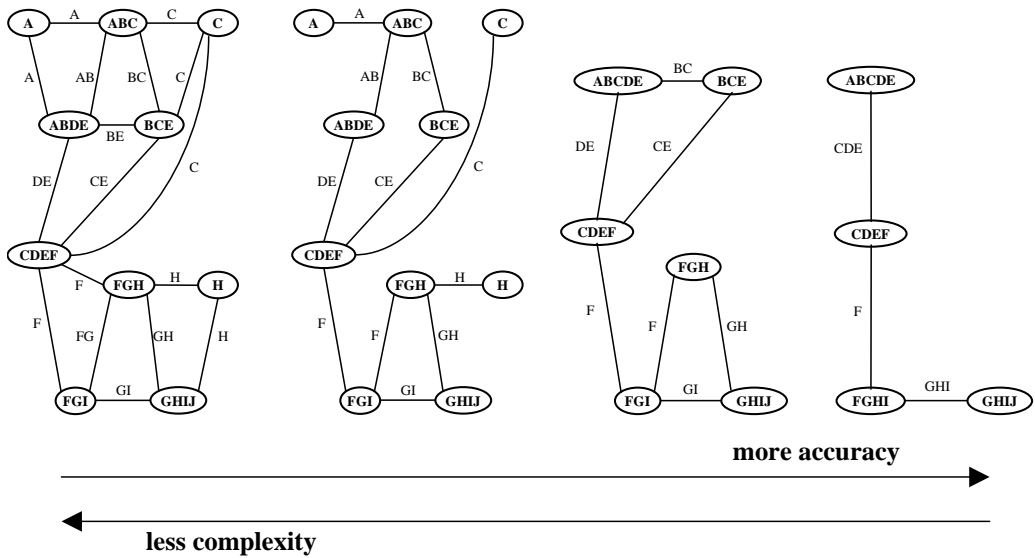

Figure 19: Join-graphs.

## 4.4 The Inference Power of IJGP

The question we address in this subsection is why propagating the messages iteratively should help. Why is IJGP upon convergence superior to IJGP with one iteration and superior to MC? One clue can be provided when considering deterministic constraint networks which can be viewed as "extreme probabilistic networks". It is known that constraint propagation algorithms, which are analogous to the messages sent by belief propagation, are guaranteed to converge and are guaranteed to improve with iteration. The propagation scheme of IJGP works similar to constraint propagation relative to the flat network abstraction of the probability distribution (where all non-zero entries are normalized to a positive constant), and propagation is guaranteed to be more accurate for that abstraction at least.

In the following we will shed some light on the IJGP's behavior by making connections with the well-known concept of arc-consistency from constraint networks (Dechter, 2003). We show that: (a) if a variable-value pair is assessed as having a zero-belief, it remains as zero-belief in subsequent iterations; (b) that any variable-value zero-beliefs computed by IJGP are correct; (c) in terms of zero/non-zero beliefs, IJGP converges in finite time. We have also empirically investigated the hypothesis that if a variable-value pair is assessed by IBP or IJGP as having a positive but very close to zero belief, then it is very likely to be correct. Although the experimental results shown in this paper do not contradict this hypothesis, some examples in more recent experiments by Dechter, Bidyuk, Mateescu, and Rollon (2010) invalidate it.

### 4.4.1 IJGP and Arc-Consistency

For any belief network we can define a constraint network that captures the assignments having strictly positive probability. We will show a correspondence between IJGP applied to the belief network and an arc-consistency algorithm applied to the constraint network. Since arc-consistency algorithms are well understood, this correspondence not only proves the target claims, but may provide additional insight into the behavior of IJGP. It justifies the iterative application of belief propagation, and it also illuminates its "distance" from being complete.

**Definition 13 (constraint satisfaction problem)** *A* Constraint Satisfaction Problem (CSP) *is a triple* $\langle X, D, C \rangle$, *where* $X = \{X_1, \ldots, X_n\}$ *is a set of variables associated with a set of discrete-valued domains* $D = \{D_1, \ldots, D_n\}$ *and a set of constraints* $C = \{C_1, \ldots, C_m\}$. *Each constraint* $C_i$ *is a pair* $\langle S_i, R_i \rangle$ *where* $R_i$ *is a relation* $R_i \subseteq D_{S_i}$ *defined on a subset of variables* $S_i \subseteq X$ *and* $D_{S_i}$ *is a Cartesian product of the domains of variables* $S_i$. *The relation* $R_i$ *denotes all compatible tuples of* $D_{S_i}$ *allowed by the constraint. A projection operator* $\pi$ *creates a new relation,* $\pi_{S_j}(R_i) = \{x | x \in D_{S_j} \text{ and } \exists y, y \in D_{S_i \setminus S_j} \text{ and } x \cup y \in R_i\}$, *where* $S_j \subseteq S_i$. *Constraints can be combined with the join operator* $\bowtie$, *resulting in a new relation,* $R_i \bowtie R_j = \{x | \pi_{S_i}(x) \in R_i \text{ and } \pi_{S_j}(x) \in R_j\}$. *A solution is an assignment of values to all the variables* $x = (x_1, \ldots, x_n), x_i \in D_i$, *such that* $\forall C_i \in C, x_{S_i} \in R_i$. *The constraint network represents its set of solutions,* $\bowtie_i C_i$.

Given a belief network $\mathcal{B}$, we define a flattening of the Bayesian network into a constraint network called $flat(\mathcal{B})$, where all the zero entries in a probability table are removed from the corresponding relation. The network $flat(\mathcal{B})$ is defined over the same set of variables and has the same set of domain values as $\mathcal{B}$.

**Definition 14 (flat network)** *Given a Bayesian network* $\mathcal{B} = \langle X, D, G, P \rangle$, *the* flat network $flat(\mathcal{B})$ *is a constraint network, where the set of variables is* $X$, *and for every* $X_i \in X$ *and its CPT* $P(X_i | pa(X_i)) \in \mathcal{B}$ *we define a constraint* $R_{F_i}$ *over the family of* $X_i$, $F_i = \{X_i\} \cup pa(X_i)$ *as follows: for every assignment* $x = (x_i, x_{pa(X_i)})$ *to* $F_i$, $(x_i, x_{pa(X_i)}) \in R_{F_i}$ *iff* $P(x_i | x_{pa(X_i)}) > 0$.

**Theorem 6** *Given a belief network* $\mathcal{B} = \langle X, D, G, P \rangle$, *where* $X = \{X_1, \ldots, X_n\}$, *for any tuple* $x = (x_1, \ldots, x_n)$: $P_{\mathcal{B}}(x) > 0 \Leftrightarrow x \in sol(flat(\mathcal{B}))$, *where* $sol(flat(\mathcal{B}))$ *is the set of solutions of the flat constraint network.*

**Proof:** $P_{\mathcal{B}}(x) > 0 \Leftrightarrow \Pi_{i=1}^n P(x_i | x_{pa(X_i)}) > 0 \Leftrightarrow \forall i \in \{1, \ldots, n\}, \ P(x_i | x_{pa(X_i)}) > 0 \Leftrightarrow \forall i \in \{1, \ldots, n\}, \ (x_i, x_{pa(X_i)}) \in R_{F_i} \Leftrightarrow x \in sol(flat(\mathcal{B}))$. □

*Constraint propagation* is a class of polynomial time algorithms that are at the center of constraint processing techniques. They were investigated extensively in the past three decades and the most well known versions are *arc-*, *path-*, and *i-consistency* (Dechter, 1992, 2003).

**Definition 15 (arc-consistency)** *(Mackworth, 1977) Given a binary constraint network* $(X, D, C)$, *the network is* arc-consistent *iff for every binary constraint* $R_{ij} \in C$, *every value* $v \in D_i$ *has a value* $u \in D_j$ *s.t.* $(v, u) \in R_{ij}$.

Note that arc-consistency is defined for binary networks, namely the relations involve at most two variables. When a binary constraint network is not arc-consistent, there are algorithms that can process it and enforce arc-consistency. The algorithms remove values from the domains of the variables that violate arc-consistency until an arc-consistent network is generated. There are several versions of improved performance arc-consistency algorithms, however we will consider a non-optimal distributed version, which we call *distributed arc-consistency*.

**DEFINITION 16 (distributed arc-consistency algorithm)** *The* algorithm distributed arc-consistency *is a message-passing algorithm over a constraint network. Each node is a variable, and maintains a current set of viable values $D_i$. Let $ne(i)$ be the set of neighbors of $X_i$ in the constraint graph. Every node $X_i$ sends a message to any node $X_j \in ne(i)$, which consists of the values in $X_j$'s domain that are consistent with the current $D_i$, relative to the constraint $R_{ji}$ that they share. Namely, the message that $X_i$ sends to $X_j$, denoted by $D_i^j$, is:*

$$D_i^j \leftarrow \pi_j(R_{ji} \bowtie D_i) \tag{1}$$

*and in addition node $i$ computes:*

$$D_i \leftarrow D_i \cap (\bowtie_{k \in ne(i)} D_k^i) \tag{2}$$

Clearly the algorithm can be synchronized into iterations, where in each iteration every node computes its current domain based on all the messages received so far from its neighbors (Eq. 2), and sends a new message to each neighbor (Eq. 1). Alternatively, Equations 1 and 2 can be combined. The message $X_i$ sends to $X_j$ is:

$$D_i^j \leftarrow \pi_j(R_{ji} \bowtie D_i \bowtie_{k \in ne(i)} D_k^i) \tag{3}$$

Next we will define a join-graph decomposition for the flat constraint network so that we can establish a correspondence between the join-graph decomposition of a Bayesian network $\mathcal{B}$ and the join-graph decomposition of its flat network $flat(\mathcal{B})$. Note that for constraint networks, the edge labeling $\theta$ can be ignored.

**DEFINITION 17 (join-graph decomposition of the flat network)** *Given a join-graph decomposition $D = \langle JG, \chi, \psi, \theta \rangle$ of a Bayesian network $\mathcal{B}$, the* join-graph decomposition $D_{flat} = \langle JG, \chi, \psi_{flat} \rangle$ *of the flat constraint network $flat(B)$ has the same underlying graph structure $JG = (V, E)$ as $D$, the same variable-labeling of the clusters $\chi$, and the mapping $\psi_{flat}$ maps each cluster to relations corresponding to CPTs, namely $R_i \in \psi_{flat}(v)$ iff CPT $p_i \in \psi(v)$.*

The distributed arc-consistency algorithm of Definition 16 can be applied to the join-graph decomposition of the flat network. In this case, the nodes that exchange messages are the clusters (namely the elements of the set $V$ of $JG$). The domain of a cluster of $V$ is the set of tuples of the join of the original relations in the cluster (namely the domain of cluster $u$ is $\bowtie_{R \in \psi_{flat}(u)} R$). The constraints are binary, and involve clusters of $V$ that are neighbors. For two clusters $u$ and $v$, their corresponding values $t_u$ and $t_v$ (which are tuples representing full assignments to the variables in the cluster) belong to the relation $R_{uv}$ (i.e., $(t_u, t_v) \in R_{u,v}$) if the projections over the separator (or labeling $\theta$) between $u$ and $v$ are identical, namely $\pi_{\theta((u,v))} t_u = \pi_{\theta((u,v))} t_v$.

We define below the algorithm *relational distributed arc-consistency* (RDAC), that applies distributed arc-consistency to any join-graph decomposition of a constraint network. We call it relational to emphasize that the nodes exchanging messages are in fact relations over the original problem variables, rather than simple variables as is the case for arc-consistency algorithms.

**DEFINITION 18 (relational distributed arc-consistency algorithm: RDAC over a join-graph)**
*Given a join-graph decomposition of a constraint network, let $R_i$ and $R_j$ be the relations of two clusters ($R_i$ and $R_j$ are the joins of the respective constraints in each cluster), having the scopes $S_i$ and $S_j$, such that $S_i \cap S_j \neq \emptyset$. The message $R_i$ sends to $R_j$ denoted $h_{(i,j)}$ is defined by:*

$$h_{(i,j)} \leftarrow \pi_{S_i \cap S_j}(R_i) \tag{4}$$

*where $ne(i) = \{j | S_i \cap S_j \neq \emptyset\}$ is the set of relations (clusters) that share a variable with $R_i$. Each cluster updates its current relation according to:*

$$R_i \leftarrow R_i \bowtie (\bowtie_{k \in ne(i)} h_{(k,i)}) \tag{5}$$

*Algorithm RDAC iterates until there is no change.*

Equations 4 and 5 can be combined, just like in Equation 3. The message that $R_i$ sends to $R_j$ becomes:

$$h_{(i,j)} \leftarrow \pi_{S_i \cap S_j}(R_i \bowtie (\bowtie_{k \in ne(i)} h_{(k,i)})) \tag{6}$$

To establish the correspondence with IJGP, we define the algorithm IJGP-RDAC that applies RDAC in the same order of computation (schedule of processing) as IJGP.

**DEFINITION 19 (IJGP-RDAC algorithm)** *Given the Bayesian network $\mathcal{B} = \langle X, D, G, P \rangle$, let $D_{flat} = \langle JG, \chi, \psi_{flat}, \theta \rangle$ be any join-graph decomposition of the flat network $flat(\mathcal{B})$. The algorithm IJGP-RDAC is applied to the decomposition $D_{flat}$ of $flat(\mathcal{B})$, and can be described as IJGP applied to D, with the following modifications:*

1. *Instead of $\prod$, we use $\bowtie$.*

2. *Instead of $\sum$, we use $\pi$.*

3. *At end end, we update the domains of variables by:*

$$D_i \leftarrow D_i \cap \pi_{X_i}((\bowtie_{v \in ne(u)} h_{(v,u)}) \bowtie (\bowtie_{R \in \psi(u)} R)) \tag{7}$$

*where $u$ is the cluster containing $X_i$.*

Note that in algorithm IJGP-RDAC, we could first merge all constraints in each cluster $u$ into a single constraint $R_u = \bowtie_{R \in \psi(u)} R$. From our construction, IJGP-RDAC enforces arc-consistency over the join-graph decomposition of the flat network. When the join-graph $D_{flat}$ is a join-tree, IJGP-RDAC solves the problem namely it finds all the solutions of the constraint network.

**Proposition 5** *Given the join-graph decomposition $D_{flat} = \langle JG, \chi, \psi_{flat}, \theta \rangle$, $JG = (V, E)$, of the flat constraint network $flat(\mathcal{B})$, corresponding to a given join-graph decomposition $D$ of a Bayesian network $\mathcal{B} = \langle X, D, G, P \rangle$, the algorithm IJGP-RDAC applied to $D_{flat}$ enforces arc-consistency over the join-graph $D_{flat}$.*

**Proof:** IJGP-RDAC applied to the join-graph decomposition $D_{flat} = \langle JG, \chi, \psi_{flat}, \theta \rangle$, $JG = (V, E)$, is equivalent to applying RDAC of Definition 18 to a constraint network that has vertices $V$ as its variables and $\{ \bowtie_{R \in \psi(u)} R | u \in V \}$ as its relations. $\quad\square$

Following the properties of convergence of arc-consistency, we can show that:

**Proposition 6** *Algorithm IJGP-RDAC converges in $O(m \cdot r)$ iterations, where $m$ is the number of edges in the join-graph and $r$ is the maximum size of a separator $D_{sep(u,v)}$ between two clusters.*

**Proof:** This follows from the fact messages (which are relations) between clusters in IJGP-RDAC change monotonically, as tuples are only successively removed from relations on separators. Since the size of each relation on a separator is bounded by $r$ and there are $m$ edges, no more than $O(m \cdot r)$ iterations will be needed. $\quad\square$

In the following we will establish an equivalence between IJGP and IJGP-RDAC in terms of zero probabilities.

**Proposition 7** *When IJGP and IJGP-RDAC are applied in the same order of computation, the messages computed by IJGP are identical to those computed by IJGP-RDAC in terms of zero / non-zero probabilities. That is, $h_{(u,v)}(x) \neq 0$ in IJGP iff $x \in h_{(u,v)}$ in IJGP-RDAC.*

**Proof:** The proof is by induction. The base case is trivially true since messages $h$ in IJGP are initialized to a uniform distribution and messages $h$ in IJGP-RDAC are initialized to complete relations.

The induction step. Suppose that $h_{(u,v)}^{IJGP}$ is the message sent from $u$ to $v$ by IJGP. We will show that if $h_{(u,v)}^{IJGP}(x) \neq 0$, then $x \in h_{(u,v)}^{IJGP-RDAC}$ where $h_{(u,v)}^{IJGP-RDAC}$ is the message sent by IJGP-RDAC from $u$ to $v$. Assume that the claim holds for all messages received by $u$ from its neighbors. Let $f \in cluster_v(u)$ in IJGP and $R_f$ be the corresponding relation in IJGP-RDAC, and $t$ be an assignment of values to variables in $elim(u, v)$. We have $h_{(u,v)}^{IJGP}(x) \neq 0 \Leftrightarrow \sum_{elim(u,v)} \prod_f f(x) \neq 0 \Leftrightarrow \exists t, \prod_f f(x, t) \neq 0 \Leftrightarrow \exists t, \forall f, f(x, t) \neq 0 \Leftrightarrow \exists t, \forall f, \pi_{scope(R_f)}(x, t) \in R_f \Leftrightarrow \exists t, \pi_{elim(u,v)}(\bowtie_{R_f} \pi_{scope(R_f)}(x, t)) \in h_{(u,v)}^{IJGP-RDAC} \Leftrightarrow x \in h_{(u,v)}^{IJGP-RDAC}$. $\quad\square$

Next we will show that IJGP computing marginal probability $P(X_i = x_i) = 0$ is equivalent to IJGP-RDAC removing $x_i$ from the domain of variable $X_i$.

**Proposition 8** *IJGP computes $P(X_i = x_i) = 0$ iff IJGP-RDAC decides that $x_i \notin D_i$.*

**Proof:** According to Proposition 7 messages computed by IJGP and IJGP-RDAC are identical in terms of zero probabilities. Let $f \in cluster(u)$ in IJGP and $R_f$ be the corresponding relation in IJGP-RDAC, and $t$ be an assignment of values to variables in $\chi(u) \backslash X_i$. We will show that when IJGP computes $P(X_i = x_i) = 0$ (upon convergence), then IJGP-RDAC computes $x_i \notin D_i$. We have $P(X_i = x_i) = \sum_{X \setminus X_i} \prod_f f(x_i) = 0 \Leftrightarrow \forall t, \prod_f f(x_i, t) = 0 \Leftrightarrow \forall t, \exists f, f(x_i, t) = 0 \Leftrightarrow \forall t, \exists R_f, \pi_{scope(R_f)}(x_i, t) \notin R_f \Leftrightarrow \forall t, (x_i, t) \notin (\bowtie_{R_f} R_f(x_i, t)) \Leftrightarrow x_i \notin D_i \cap \pi_{X_i}(\bowtie_{R_f} R_f(x_i, t)) \Leftrightarrow x_i \notin D_i$. Since arc-consistency is sound, so is the decision of zero probabilities. $\square$

Next we will show that $P(X_i = x_i) = 0$ computed by IJGP is sound.

**Theorem 7** *Whenever IJGP finds $P(X_i = x_i) = 0$, then the probability $P(X_i)$ expressed by the Bayesian network conditioned on the evidence is 0 as well.*

**Proof:** According to Proposition 8, whenever IJGP finds $P(X_i = x_i) = 0$, the value $x_i$ is removed from the domain $D_i$ by IJGP-RDAC, therefore value $x_i \in D_i$ is a no-good of the network $flat(\mathcal{B})$, and from Theorem 6 it follows that $P_{\mathcal{B}}(X_i = x_i) = 0$. $\square$

In the following we will show that the time it takes IJGP to find all $P(X_i = x_i) = 0$ is bounded.

**Proposition 9** *IJGP finds all $P(X_i = x_i) = 0$ in finite time, that is, there exists a number $k$, such that no $P(X_i = x_i) = 0$ will be found after $k$ iterations.*

**Proof:** This follows from the fact that the number of iterations it takes for IJGP to compute $P(X_i = x_i) = 0$ is exactly the same number of iterations IJGP-RDAC takes to remove $x_i$ from the domain $D_i$ (Proposition 7 and Proposition 8), and the fact the IJGP-RDAC runtime is bounded (Proposition 6). $\square$

Previous results also imply that IJGP is monotonic with respect to zeros.

**Proposition 10** *Whenever IJGP finds $P(X_i = x_i) = 0$, it stays 0 during all subsequent iterations.*

**Proof:** Since we know that relations in IJGP-RDAC are monotonically decreasing as the algorithm progresses, it follows from the equivalence of IJGP-RDAC and IJGP (Proposition 7) that IJGP is monotonic with respect to zeros. $\square$

### 4.4.2 A Finite Precision Problem

On finite precision machines there is the danger that an underflow can be interpreted as a zero value. We provide here a warning that an implementation of belief propagation should not allow the creation of zero values by underflow. We show an example in Figure 20 where IBP's messages converge in the limit (i.e., in an infinite number of iterations), but they do not stabilize in any finite number of iterations. If all the nodes $H_k$ are set to value 1, the belief for any of the $X_i$ variables as a function of iteration is given in the table in Figure 20. After about 300 iterations, the finite precision of our computer is not able to represent the value for $Bel(X_i = 3)$, and this appears to be zero, yielding the final updated belief $(.5, .5, 0)$, when in fact the true updated belief should be $(0, 0, 1)$. Notice that $(.5, .5, 0)$ cannot be regarded as a legitimate fixed point for IBP. Namely, if we would initialize IBP with the values $(.5, .5, 0)$, then the algorithm would maintain them, appearing to have a fixed point, but initializing IBP with zero values cannot be expected to be correct. When we initialize with zeros we forcibly introduce determinism in the model, and IBP will always maintain it afterwards.

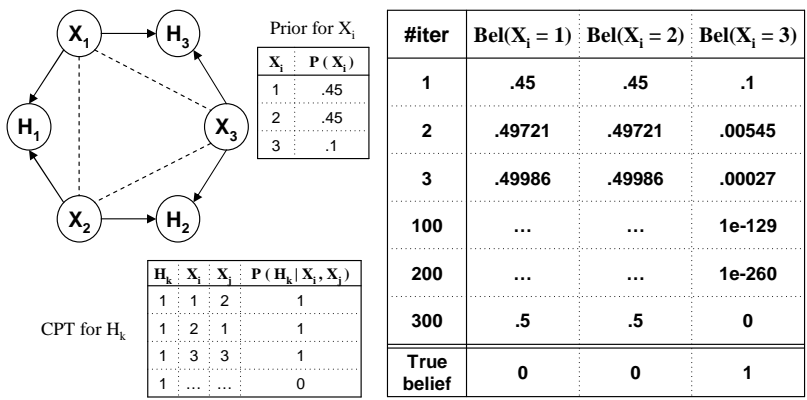

Figure 20: Example of a finite precision problem.

However, this example does not contradict our theory because, mathematically, $Bel(X_i = 3)$ never becomes a true zero, and IBP never reaches a quiescent state. The example shows that a close to zero belief network can be arbitrarily inaccurate. In this case the inaccuracy seems to be due to the initial prior belief which are so different from the posterior ones.

### 4.4.3 ACCURACY OF IBP ACROSS BELIEF DISTRIBUTION

We present an empirical evaluation of the accuracy of IBP's prediction for the range of belief distribution from 0 to 1. These results also extend to IJGP. In the previous section, we proved that zero values inferred by IBP are correct, and we wanted to test the hypothesis that this property extends to $\epsilon$ small beliefs (namely, that are very close to zero). That is, if IBP infers a posterior belief close to zero, then it is likely to be correct. The results presented in this paper seem to support the hypothesis, however new experiments by Dechter et al. (2010) show that it is not true in general. We do not have yet a good characterization of the cases when the hypothesis is confirmed.

To test this hypothesis, we computed the absolute error of IBP per intervals of $[0, 1]$. For a given interval $[a, b]$, where $0 \le a < b \le 1$, we use measures inspired from information retrieval: *Recall Absolute Error* and *Precision Absolute Error*.

*Recall* is the absolute error averaged over all the exact posterior beliefs that fall into the interval $[a, b]$. For *Precision*, the average is taken over all the approximate posterior belief values computed by IBP to be in the interval $[a, b]$. Intuitively, *Recall([a,b])* indicates how far the belief computed by IBP is from the exact, when the exact is in $[a, b]$; *Precision([a,b])* indicates how far the exact is from IBP's prediction, when the value computed by IBP is in $[a, b]$.

Our experiments show that the two measures are strongly correlated. We also show the histograms of distribution of belief for each interval, for the exact and for IBP, which are also strongly correlated. The results are given in Figures 21 and 22. The left Y axis corresponds to the histograms (the bars), the right Y axis corresponds to the absolute error (the lines).

We present results for two classes of problems: coding networks and grid network. All problems have binary variables, so the graphs are symmetric about 0.5 and we only show the interval $[0, 0.5]$. The number of variables, number of iterations and induced width w* are reported for each graph.

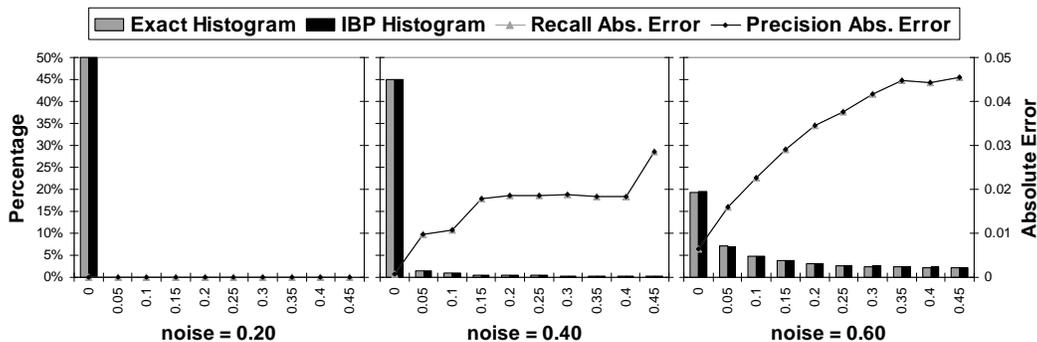

Figure 21: Coding, N=200, 1000 instances, w*=15.

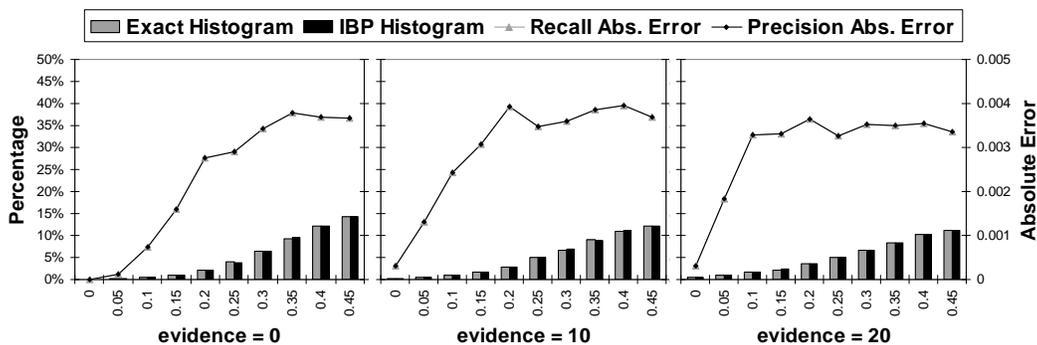

Figure 22: 10x10 grids, 100 instances, w*=15.

**Coding networks** IBP is famously known to have impressive performance on coding networks. We tested on linear block codes, with 50 nodes per layer and 3 parent nodes. Figure 21 shows the results for three different values of channel noise: 0.2, 0.4 and 0.6. For noise 0.2, all the beliefs computed by IBP are extreme. The Recall and Precision are very small, of the order of $10^{-11}$. So, in this case, all the beliefs are very small ($\epsilon$ small) and IBP is able to infer them correctly, resulting in almost perfect accuracy (IBP is indeed perfect in this case for the bit error rate). When the noise is increased, the Recall and Precision tend to get closer to a bell shape, indicating higher error for values close to 0.5 and smaller error for extreme values. The histograms also show that less belief values are extreme as the noise is increased, so all these factors account for an overall decrease in accuracy as the channel noise increases. These networks are examples with a large number of $\epsilon$-small probabilities and IBP is able to infer them correctly (absolute error is small).

**Grid networks** We present results for grid networks in Figure 22. Contrary to the case of coding networks, the histograms show higher concentration of beliefs around 0.5. However, the accuracy is still very good for beliefs close to zero. The absolute error peaks close to 0 and maintains a plateau, as evidence is increased, indicating less accuracy for IBP.

## 5. Experimental Evaluation

As we anticipated in the summary of Section 3, and as can be clearly seen now by the structuring of a bounded join-graph, there is a close relationship between the mini-clustering algorithm MC(i)

| | | Absolute error | | | | Relative error | | | | KL distance | | | | Time | | | |
|---|---|---|---|---|---|---|---|---|---|---|---|---|---|---|---|---|---|
| | | IBP | IJGP | | | IBP | IJGP | | | IBP | IJGP | | | IBP | IJGP | | |
| #it | #evid | | i=2 | i=5 | i=8 | | i=2 | i=5 | i=8 | | i=2 | i=5 | i=8 | | i=2 | i=5 | i=8 |
| 1 | 0 | 0.02988 | 0.03055 | 0.02623 | 0.02940 | 0.06388 | 0.15694 | 0.05677 | 0.07153 | 0.00213 | 0.00391 | 0.00208 | 0.00277 | 0.0017 | 0.0036 | 0.0058 | 0.0295 |
| | 5 | 0.06178 | 0.04434 | 0.04201 | 0.04554 | 0.15005 | 0.12340 | 0.12056 | 0.11154 | 0.00812 | 0.00582 | 0.00478 | 0.00558 | 0.0013 | 0.0040 | 0.0052 | 0.0200 |
| | 10 | 0.08762 | 0.05777 | 0.05409 | 0.05910 | 0.23777 | 0.18071 | 0.14278 | 0.15686 | 0.01547 | 0.00915 | 0.00768 | 0.00899 | 0.0013 | 0.0040 | 0.0036 | 0.0121 |
| 5 | 0 | 0.00829 | 0.00636 | 0.00592 | 0.00669 | 0.01726 | 0.01326 | 0.01239 | 0.01398 | 0.00021 | 0.00014 | 0.00015 | 0.00018 | 0.0066 | 0.0145 | 0.0226 | 0.1219 |
| | 5 | 0.05182 | 0.00886 | 0.00886 | 0.01123 | 0.12589 | 0.01967 | 0.01965 | 0.02494 | 0.00658 | 0.00024 | 0.00026 | 0.00044 | 0.0060 | 0.0120 | 0.0185 | 0.0840 |
| | 10 | 0.08039 | 0.01155 | 0.01073 | 0.01399 | 0.21781 | 0.03014 | 0.02553 | 0.03279 | 0.01382 | 0.00055 | 0.00042 | 0.00073 | 0.0048 | 0.0100 | 0.0138 | 0.0536 |
| 10 | 0 | 0.00828 | 0.00584 | 0.00514 | 0.00495 | 0.01725 | 0.01216 | 0.01069 | 0.01030 | 0.00021 | 0.00012 | 0.00010 | 0.00010 | 0.0130 | 0.0254 | 0.0436 | 0.2383 |
| | 5 | 0.05182 | 0.00774 | 0.00732 | 0.00708 | 0.12590 | 0.01727 | 0.01618 | 0.01566 | 0.00658 | 0.00018 | 0.00017 | 0.00016 | 0.0121 | 0.0223 | 0.0355 | 0.1639 |
| | 10 | 0.08040 | 0.00892 | 0.00808 | 0.00855 | 0.21782 | 0.02101 | 0.01907 | 0.02005 | 0.01382 | 0.00028 | 0.00024 | 0.00029 | 0.0109 | 0.0191 | 0.0271 | 0.1062 |
| MC | 0 | | 0.04044 | 0.04287 | 0.03748 | | 0.08811 | 0.09342 | 0.08117 | | 0.00403 | 0.00435 | 0.00369 | | 0.0159 | 0.0173 | 0.0552 |
| | 5 | | 0.05303 | 0.05171 | 0.04250 | | 0.12375 | 0.11775 | 0.09596 | | 0.00659 | 0.00636 | 0.00477 | | 0.0146 | 0.0158 | 0.0532 |
| | 10 | | 0.06033 | 0.05489 | 0.04266 | | 0.14702 | 0.13219 | 0.10074 | | 0.00841 | 0.00729 | 0.00503 | | 0.0119 | 0.0143 | 0.0470 |

Table 4: Random networks: N=50, K=2, C=45, P=3, 100 instances, w*=16.

and IJGP(i). In particular, one iteration of IJGP(i) is similar to MC(i). MC sends messages up and down along the clusters that form a set of trees. IJGP has additional connections that allow more interaction between the mini-clusters of the same cluster. Since this is a cyclic structure, iterating is facilitated, with its virtues and drawbacks.s

In our evaluation of IJGP(i), we focus on two different aspects: (a) the sensitivity of parametric IJGP(i) to its i-bound and to the number of iterations; (b) a comparison of IJGP(i) with publicly available state-of-the-art approximation schemes.

## 5.1 Effect of i-bound and Number of Iterations

We tested the performance of IJGP(i) on random networks, on M-by-M grids, on the two benchmark CPCS files with 54 and 360 variables, respectively and on coding networks. On each type of networks, we ran IBP, MC(i) and IJGP(i), while giving IBP and IJGP(i) the same number of iterations.

We use the partitioning method described in Section 4.3 to construct a join-graph. To determine the order of message computation, we recursively pick an edge (u,v), such that node u has the fewest incoming messages missing.

For each network except coding, we compute the exact solution and compare the accuracy using the absolute and relative error, as before, as well as the KL (Kullback-Leibler) distance - $P_{exact}(X = a) \cdot log(P_{exact}(X = a)/P_{approximation}(X = a))$ averaged over all values, all variables and all problems. For coding networks we report the Bit Error Rate (BER) computed as described in Section 3.2. We also report the time taken by each algorithm.

The random networks were generated using parameters (N,K,C,P), where N is the number of variables, K is their domain size, C is the number of conditional probability tables (CPTs) and P is the number of parents in each CPT. Parents in each CPT are picked randomly and each CPT is filled randomly. In grid networks, N is a square number and each CPT is filled randomly. In each problem class, we also tested different numbers of evidence variables. As before, the coding networks are from the class of linear block codes, where $\sigma$ is the channel noise level. Note that we are limited to relatively small and sparse problem instances because our evaluation measures are based on comparing against exact figures.

**Random networks** results for networks having N=50, K=2, C=45 and P=3 are given in Table 4 and in Figures 23 and 24. For IJGP(i) and MC(i) we report 3 different values of i-bound: 2, 5, 8. For IBP and IJGP(i) we report results for 3 different numbers of iterations: 1, 5, 10. We report results for 3 different numbers of evidence: 0, 5, 10. From Table 4 and Figure 23a we see that IJGP(i) is always better than IBP (except when i=2 and number of iterations is 1), sometimes by an order of magnitude, in terms of absolute error, relative error and KL distance. IBP rarely changes after 5 iterations, whereas IJGP(i)'s solution can be improved with more iterations (up to 15-20). As theory predicted, the accuracy of IJGP(i) for one iteration is about the same as that of MC(i). But IJGP(i) improves as the number of iterations increases, and is eventually better than MC(i) by as much as an order of magnitude, although it clearly takes more time, especially when the i-bound is large.

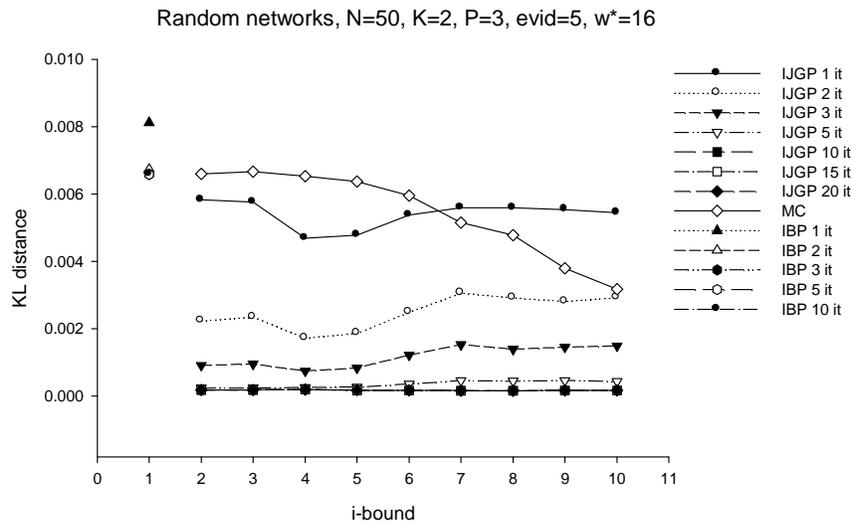

(a) Performance vs. i-bound.

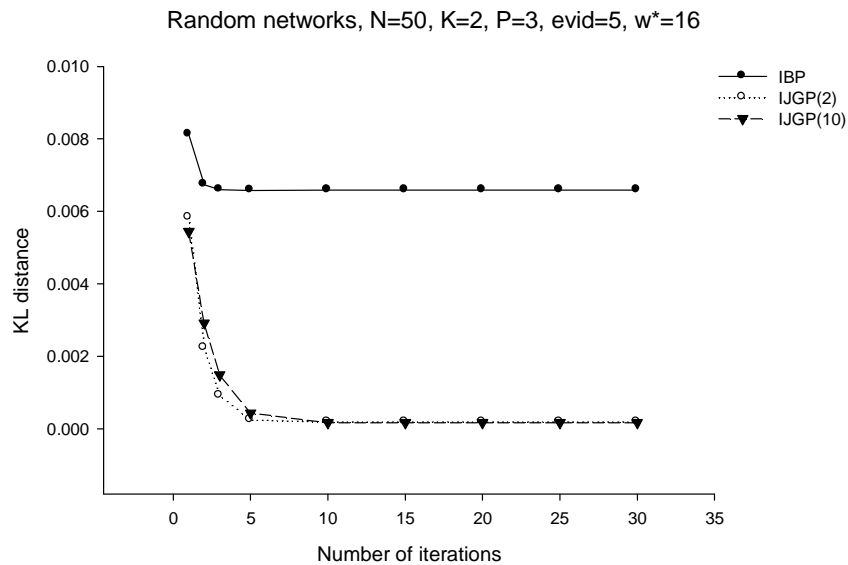

(b) Convergence with iterations.

Figure 23: Random networks: KL distance.

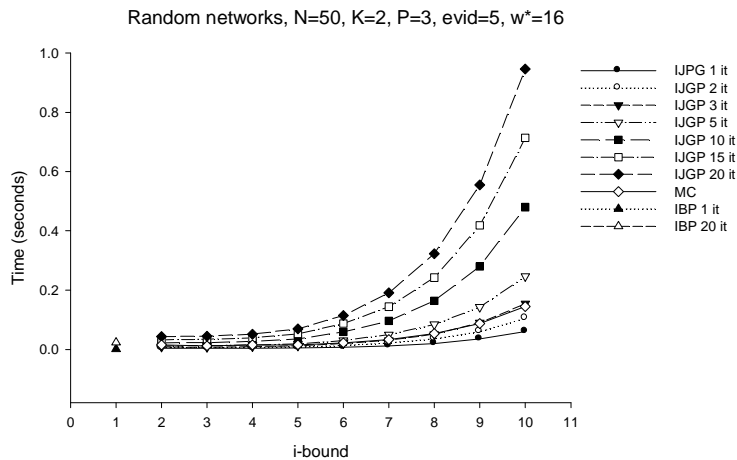

Figure 24: Random networks: Time.

| | | Absolute error | | | | Relative error | | | | KL distance | | | | Time | | | |
|---|---|---|---|---|---|---|---|---|---|---|---|---|---|---|---|---|---|
| | | IBP | IJGP | | | IBP | IJGP | | | IBP | IJGP | | | IBP | IJGP | | |
| #it | #evid | | i=2 | i=5 | i=8 | | i=2 | i=5 | i=8 | | i=2 | i=5 | i=8 | | i=2 | i=5 | i=8 |
| 1 | 0 | 0.03524 | 0.05550 | 0.04292 | 0.03318 | 0.08075 | 0.13533 | 0.10252 | 0.07904 | 0.00289 | 0.00859 | 0.00602 | 0.00454 | 0.0010 | 0.0053 | 0.0106 | 0.0426 |
| | 5 | 0.05375 | 0.05284 | 0.04012 | 0.03661 | 0.16380 | 0.13225 | 0.09889 | 0.09116 | 0.00725 | 0.00570 | 0.00549 | 0.00454 | 0.0016 | 0.0041 | 0.0092 | 0.0315 |
| | 10 | 0.07094 | 0.05453 | 0.04304 | 0.03966 | 0.23624 | 0.14588 | 0.12492 | 0.12202 | 0.01232 | 0.00905 | 0.00681 | 0.00653 | 0.0013 | 0.0038 | 0.0072 | 0.0256 |
| 5 | 0 | 0.00358 | 0.00393 | 0.00325 | 0.00284 | 0.00775 | 0.00849 | 0.00702 | 0.00634 | 0.00005 | 0.00006 | 0.00007 | 0.00010 | 0.0049 | 0.0152 | 0.0347 | 0.1462 |
| | 5 | 0.03224 | 0.00379 | 0.00319 | 0.00296 | 0.11299 | 0.00844 | 0.00710 | 0.00669 | 0.00483 | 0.00006 | 0.00007 | 0.00010 | 0.0053 | 0.0131 | 0.0309 | 0.1127 |
| | 10 | 0.05503 | 0.00364 | 0.00316 | 0.00314 | 0.19403 | 0.00841 | 0.00756 | 0.01313 | 0.00994 | 0.00006 | 0.00009 | 0.00019 | 0.0036 | 0.0127 | 0.0271 | 0.0913 |
| 10 | 0 | 0.00352 | 0.00352 | 0.00232 | 0.00136 | 0.00760 | 0.00760 | 0.00502 | 0.00293 | 0.00005 | 0.00005 | 0.00003 | 0.00001 | 0.0090 | 0.0277 | 0.0671 | 0.2776 |
| | 5 | 0.03222 | 0.00357 | 0.00248 | 0.00149 | 0.11295 | 0.00796 | 0.00549 | 0.00330 | 0.00483 | 0.00005 | 0.00003 | 0.00002 | 0.0096 | 0.0246 | 0.0558 | 0.2149 |
| | 10 | 0.05503 | 0.00347 | 0.00239 | 0.00141 | 0.19401 | 0.00804 | 0.00556 | 0.00328 | 0.00994 | 0.00005 | 0.00003 | 0.00001 | 0.0090 | 0.0223 | 0.0495 | 0.1716 |
| MC | 0 | | 0.05827 | 0.04036 | 0.01579 | | 0.13204 | 0.08833 | 0.03440 | | 0.00650 | 0.00387 | 0.00105 | | 0.0106 | 0.0142 | 0.0382 |
| | 5 | | 0.05973 | 0.03692 | 0.01355 | | 0.13831 | 0.08213 | 0.03001 | | 0.00696 | 0.00348 | 0.00099 | | 0.0102 | 0.0130 | 0.0342 |
| | 10 | | 0.05866 | 0.03416 | 0.01075 | | 0.14120 | 0.07791 | 0.02488 | | 0.00694 | 0.00326 | 0.00075 | | 0.0099 | 0.0116 | 0.0321 |

Table 5: 9x9 grid, K=2, 100 instances, w*=12.

Figure 23a shows a comparison of all algorithms with different numbers of iterations, using the KL distance. Because the network structure changes with different i-bounds, we do not necessarily see monotonic improvement of IJGP with i-bound for a given number of iterations (as is the case with MC). Figure 23b shows how IJGP converges with more iterations to a smaller KL distance than IBP. As expected, the time taken by IJGP (and MC) varies exponentially with the i-bound (see Figure 24).

**Grid networks** results with networks of N=81, K=2, 100 instances are very similar to those of random networks. They are reported in Table 5 and in Figure 25, where we can see the impact of having evidence (0 and 5 evidence variables) on the algorithms. IJGP at convergence gives the best performance in both cases, while IBP's performance deteriorates with more evidence and is surpassed by MC with i-bound 5 or larger.

**CPCS networks** results with CPCS54 and CPCS360 are given in Table 6 and Figure 26, and are even more pronounced than those of random and grid networks. When evidence is added, IJGP(i) is more accurate than MC(i), which is more accurate than IBP, as can be seen in Figure 26a.

**Coding networks** results are given in Table 7. We tested on large networks of 400 variables, with treewidth w*=43, with IJGP and IBP set to run 30 iterations (this is more than enough to ensure convergence). IBP is known to be very accurate for this class of problems and it is indeed better than MC. However we notice that IJGP converges to slightly smaller BER than IBP even for small values of the i-bound. Both the coding network and CPCS360 show the scalability of IJGP for large size problems. Notice that here the anytime behavior of IJGP is not clear.

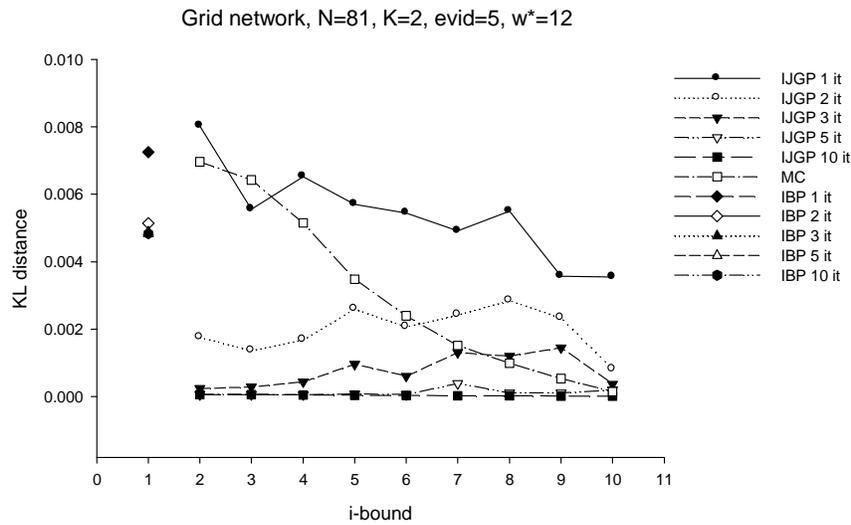

(a) Performance vs. i-bound.

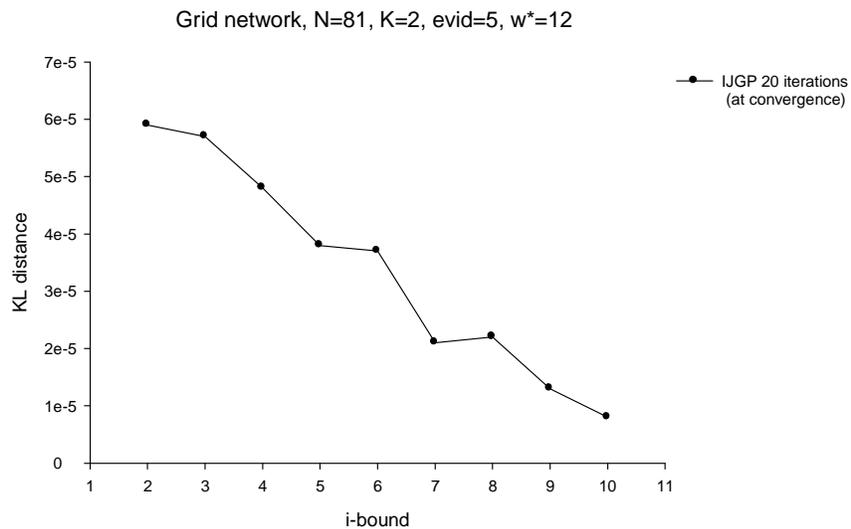

(b) Fine granularity for KL.

Figure 25: Grid 9x9: KL distance.

In summary, we see that IJGP is almost always superior to both IBP and MC(i) and is sometimes more accurate by several orders of magnitude. One should note that IBP cannot be improved with more time, while MC(i) requires a large i-bound for many hard and large networks to achieve reasonable accuracy. There is no question that the iterative application of IJGP is instrumental to its success. In fact, IJGP(2) in isolation appears to be the most cost-effective variant.

| #it | #evid | Absolute error | | | | Relative error | | | | KL distance | | | | Time | | | |
|---|---|---|---|---|---|---|---|---|---|---|---|---|---|---|---|---|---|
| | | IBP | IJGP | | | IBP | IJGP | | | IBP | IJGP | | | IBP | IJGP | | |
| | | | i=2 | i=5 | i=8 | | i=2 | i=5 | i=8 | | i=2 | i=5 | i=8 | | i=2 | i=5 | i=8 |
| | | | | | | | CPCS54 | | | | | | | | | | |
| 1 | 0 | 0.01324 | 0.03747 | 0.03183 | 0.02233 | 0.02716 | 0.08966 | 0.07761 | 0.05616 | 0.00041 | 0.00583 | 0.00512 | 0.00378 | 0.0097 | 0.0137 | 0.0146 | 0.0275 |
| | 5 | 0.02684 | 0.03739 | 0.03124 | 0.02337 | 0.05736 | 0.09007 | 0.07676 | 0.05856 | 0.00199 | 0.00573 | 0.00493 | 0.00366 | 0.0072 | 0.0094 | 0.0087 | 0.0169 |
| | 10 | 0.03915 | 0.03843 | 0.03426 | 0.02747 | 0.08475 | 0.09156 | 0.08246 | 0.06687 | 0.00357 | 0.00567 | 0.00506 | 0.00390 | 0.005 | 0.0047 | 0.0052 | 0.0115 |
| 5 | 0 | 0.00031 | 0.00016 | 0.00123 | 0.00110 | 0.00064 | 0.00033 | 0.00255 | 0.00225 | 7.75e-7 | 0.00000 | 0.00002 | 0.00001 | 0.0189 | 0.0371 | 0.0334 | 0.0912 |
| | 5 | 0.01874 | 0.00058 | 0.00092 | 0.00098 | 0.00461 | 0.00124 | 0.00194 | 0.00203 | 0.00161 | 0.00000 | 0.00001 | 0.00001 | 0.0337 | 0.0215 | 0.0260 | 0.0631 |
| | 10 | 0.03348 | 0.00010 | 0.00139 | 0.00144 | 0.07302 | 0.00215 | 0.00298 | 0.00382 | 0.00321 | 0.00001 | 0.00003 | 0.00001 | 0.0290 | 0.0144 | 0.0178 | 0.0378 |
| 10 | 0 | 0.00031 | 0.00009 | 0.00014 | 0.00015 | 0.00064 | 0.00018 | 0.00029 | 0.00031 | 7.75e-7 | 0.00000 | 0.00000 | 0.00000 | 0.0736 | 0.0587 | 0.0667 | 0.1720 |
| | 5 | 0.01874 | 0.00037 | 0.00034 | 0.00038 | 0.04067 | 0.00078 | 0.00071 | 0.00080 | 0.00161 | 0.00000 | 0.00000 | 0.00000 | 0.0633 | 0.0389 | 0.0471 | 0.1178 |
| | 10 | 0.03348 | 0.00058 | 0.00051 | 0.00057 | 0.07302 | 0.00123 | 0.00109 | 0.00122 | 0.00321 | 4.0e-6 | 3.0e-6 | 4.0e-6 | 0.0575 | 0.0251 | 0.0297 | 0.0723 |
| MC | 0 | | 0.02721 | 0.02487 | 0.01486 | | 0.05648 | 0.05128 | 0.03047 | | 0.00218 | 0.00171 | 0.00076 | | 0.0144 | 0.0125 | 0.0333 |
| | 5 | | 0.02702 | 0.02522 | 0.01760 | | 0.05687 | 0.05314 | 0.03713 | | 0.00202 | 0.00186 | 0.00098 | | 0.0103 | 0.0126 | 0.0346 |
| | 10 | | 0.02825 | 0.02504 | 0.01600 | | 0.06002 | 0.05318 | 0.03409 | | 0.00216 | 0.00177 | 0.00091 | | 0.0094 | 0.0090 | 0.0295 |
| | | | | | | | CPCS360 | | | | | | | | | | |
| 1 | 10 | 0.26421 | 0.14222 | 0.13907 | 0.14334 | 7.78167 | 2119.20 | 2132.78 | 2133.84 | 0.17974 | 0.09297 | 0.09151 | 0.09255 | 0.7172 | 0.5486 | 0.5282 | 0.4593 |
| | 20 | 0.26326 | 0.12867 | 0.12937 | 0.13665 | 370.444 | 28720.38 | 30704.93 | 31689.59 | 0.17845 | 0.08212 | 0.08269 | 0.08568 | 0.6794 | 0.5547 | 0.5250 | 0.4578 |
| 10 | 10 | 0.01772 | 0.00694 | 0.00121 | 0.00258 | 1.06933 | 6.07399 | 0.01005 | 0.00330 | 0.017718 | 0.00203 | 0.00019 | 0.00116 | 7.2205 | 4.7781 | 4.5191 | 3.7906 |
| | 20 | 0.02413 | 0.00466 | 0.00115 | 0.00138 | 62.99310 | 26.04308 | 0.00886 | 0.01353 | 0.02027 | 0.00118 | 0.00015 | 0.00036 | 7.0830 | 4.8705 | 4.6468 | 3.8392 |
| 20 | 10 | 0.01772 | 0.00003 | 3.0e-6 | 3.0e-6 | 1.06933 | 0.00044 | 8.0e-6 | 7.0e-6 | 0.01771 | 5.0e-6 | 0.0 | 0.0 | 14.4379 | 9.5783 | 9.0770 | 7.6017 |
| | 20 | 0.02413 | 0.00001 | 9.0e-6 | 9.0e-6 | 62.9931 | 0.00014 | 0.00013 | 0.00004 | 0.02027 | 0.0 | 0.0 | 0.0 | 13.6064 | 9.4582 | 9.0423 | 7.4453 |
| MC | 10 | | 0.03389 | 0.01984 | 0.01402 | | 0.65600 | 0.20023 | 0.11990 | | 0.01299 | 0.00590 | 0.00390 | | 2.8077 | 2.7112 | 2.5188 |
| | 20 | | 0.02715 | 0.01543 | 0.00957 | | 0.81401 | 0.17345 | 0.09113 | | 0.01007 | 0.00444 | 0.00234 | | 2.8532 | 2.7032 | 2.5297 |

Table 6: CPCS54 50 instances, w*=15; CPCS360 10 instances, w*=20.

| | | Bit Error Rate | | | | | |
|---|---|---|---|---|---|---|---|
| | | i-bound | | | | | |
| σ | | 2 | 4 | 6 | 8 | 10 | IBP |
| 0.22 | IJGP | 0.00005 | 0.00005 | 0.00005 | 0.00005 | 0.00005 | 0.00005 |
| | MC | 0.00501 | 0.00800 | 0.00586 | 0.00462 | 0.00392 | |
| 0.28 | IJGP | 0.00062 | 0.00062 | 0.00062 | 0.00062 | 0.00062 | 0.00064 |
| | MC | 0.02170 | 0.02968 | 0.02492 | 0.02048 | 0.01840 | |
| 0.32 | IJGP | 0.00238 | 0.00238 | 0.00238 | 0.00238 | 0.00238 | 0.00242 |
| | MC | 0.04018 | 0.05004 | 0.04480 | 0.03878 | 0.03558 | |
| 0.40 | IJGP | 0.01202 | 0.01188 | 0.01194 | 0.01210 | 0.01192 | 0.01220 |
| | MC | 0.08726 | 0.09762 | 0.09272 | 0.08766 | 0.08334 | |
| 0.51 | IJGP | 0.07664 | 0.07498 | 0.07524 | 0.07578 | 0.07554 | 0.07816 |
| | MC | 0.15396 | 0.16048 | 0.15710 | 0.15452 | 0.15180 | |
| 0.65 | IJGP | 0.19070 | 0.19056 | 0.19016 | 0.19030 | 0.19056 | 0.19142 |
| | MC | 0.21890 | 0.22056 | 0.21928 | 0.21904 | 0.21830 | |
| | | Time | | | | | |
| | IJGP | 0.36262 | 0.41695 | 0.86213 | 2.62307 | 9.23610 | 0.019752 |
| | MC | 0.25281 | 0.21816 | 0.31094 | 0.74851 | 2.33257 | |

Table 7: Coding networks: N=400, P=4, 500 instances, 30 iterations, w*=43.

## 5.2 Comparing IJGP with Other Algorithms

In this section we provide a comparison of IJGP with state-of-the-art publicly available schemes. The comparison is based on a recent evaluation of algorithms performed at the Uncertainty in AI 2008 conference[4]. We will present results on solving the belief updating task (also called the task of computing posterior node marginals - MAR). We first give a brief overview of the schemes that we experimented and compared with.

1. EDBP - Edge Deletion for Belief Propagation

---

4. Complete results are available at http://graphmod.ics.uci.edu/uai08/Evaluation/Report.

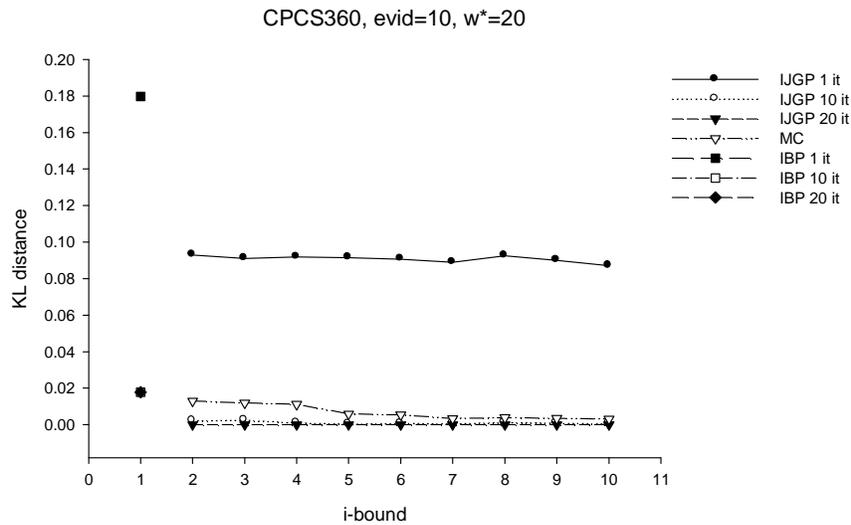

(a) Performance vs. i-bound.

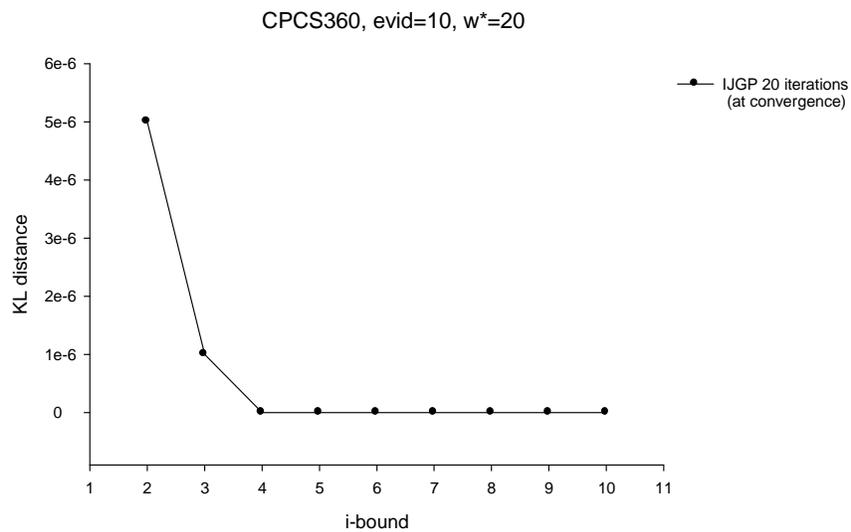

(b) Fine granularity for KL.

Figure 26: CPCS360: KL distance.

EDBP (Choi & Darwiche, 2006a, 2006b) is an approximation algorithm for Belief Updating. It solves exactly a simplified version of the original problem, obtained by deleting some of the edges of the problem graph. Edges to be deleted are selected based on two criteria: quality of approximation and complexity of computation (tree-width reduction). Information loss from lost dependencies is compensated for by introducing auxiliary network parameters. This method corresponds to Iterative Belief Propagation (IBP) when enough edges are deleted to yield a poly-tree, and corresponds to generalized BP otherwise.

2. TLSBP - A truncated Loop series Belief propagation algorithm

TLSBP is based on the loop series expansion formula of Chertkov and Chernyak (2006) which specifies a series of terms that need to be added to the solution output by BP so that the exact solution can be recovered. This series is basically a sum over all so-called generalized loops in the graph. Unfortunately, because the number of these generalized loops can be prohibitively large, the series is of little value. The idea in TLSBP is to truncate the series by decomposing all generalized loops into simple and smaller loops, thus limiting the number of loops to be summed. In our evaluation, we used an implementation of TLSBP available from the work of Gomez, Mooji, and Kappen (2007). The implementation can handle binary networks only.

### 3. EPIS - Evidence Pre-propagation Importance Sampling

EPIS (Yuan & Druzdzel, 2003) is an importance sampling algorithm for Belief Updating. It is well known that sampling algorithms perform poorly when presented with unlikely evidence. However, when samples are weighted by an importance function, good approximation can be obtained. This algorithm computes an approximate importance function using loopy belief propagation and $\epsilon$-cutoff heuristic. We used an implementation of EPIS available from the authors. The implementation works on Bayesian networks only.

### 4. IJGP - Iterative Join-Graph Propagation

In the evaluation, IJGP(i) was first run with $i=2$, until convergence, then with $i=3$, until convergence, etc. until $i=$ treewidth (when $i$-bound=treewidth, the join-graph becomes a join-tree and IJGP becomes exact). As preprocessing, the algorithm performed SAT-based variable domain pruning by converting zero probabilities in the problem to a SAT problem and performing singleton-consistency enforcement. Because the problem size may reduce substantially, in some cases, this preprocessing step may have a significant impact on the time-complexity of IJGP, amortized over the increasing $i$-bound. However, for a given $i$-bound, this step improves the accuracy of IJGP only marginally.

### 5. SampleSearch

SampleSearch (Gogate & Dechter, 2007) is a specialized importance sampling scheme for graphical models that contain zero probabilities in their CPTs. On such graphical models, importance sampling suffers from the rejection problem in that it generates a large number of samples which have zero weight. SampleSearch circumvents the rejection problem by sampling from the backtrack-free search space in which every assignment (sample) is guaranteed to have non-zero weight. The backtrack-free search space is constructed on the fly by interleaving sampling with backtracking style search. Namely, when a sample is supposed to be rejected because its weight is zero, the algorithm continues instead with systematic backtracking search, until a non zero weight sample is found. For the evaluation version, the importance distribution of SampleSearch was constructed from the output of IJGP with $i$-bound of 3. For more information on how the importance distribution is constructed from the output of IJGP, see the work by Gogate (2009).

The evaluation was conducted on the following benchmarks (see footnote 4 for details):

1. UAI06-MPE - from UAI-06, 57 instances, Bayesian networks (40 instances were used).

2. UAI06-PE - from UAI-06, 78 instances, Bayesian networks (58 instances were used).

|              | WCSPs | BN2O | Grids | Linkage | Promedas | UAI06-MPE | UAI06-PE | Relational |
|--------------|:-----:|:----:|:-----:|:-------:|:--------:|:---------:|:--------:|:----------:|
| IJGP         | √     | √    | √     | √       | √        | √         | √        | √          |
| EDBP         | √     | √    | √     | √       | √        | √         | √        | √          |
| TLSBP        |       | √    | √     |         | √        |           |          | √          |
| EPIS         |       | √    | √     |         |          | √         | √        | √          |
| SampleSearch | √     | √    | √     | √       | √        | √         | √        | √          |

Table 8: Scope of our experimental study.

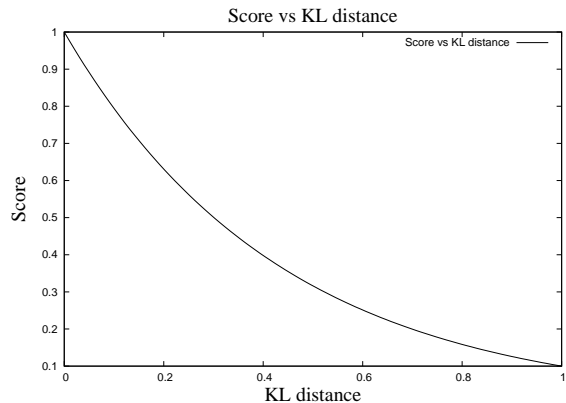

Figure 27: Score as a function of KL distance.

3. Relational Bayesian networks - constructed from the Primula tool, 251 instances, binary variables, large networks with large tree-width, but with high levels of determinism (30 instances were used).

4. Linkage networks - 22 instances, tree-width 20-35, Markov networks (5 instances were used).

5. Grids - from 12x12 to 50x50, 320 instances, treewidth 12-50.

6. BN2O networks - Two-layer Noisy-OR Bayesian networks, 18 instances, binary variables, up to 55 variables, treewidth 24-27.

7. WCSPs - Weighted CSPs, 97 instances, Markov networks (18 instances were used).

8. Promedas - real-world medical diagnosis, 238 instances, tree-width 1-60, Markov networks (46 instances were used).

Table 8 shows the scope of our experimental study. A √ indicates that the solver was able to handle the benchmark type and therefore evaluated on it while a lack of a √ indicates otherwise.

We measure the performance of the algorithms in terms of a KL-distance based score. Formally, the score of a solver on a problem instance is equal to $10^{-avgkld}$ where avgkld is the average KL distance between the exact marginal (which was computed using the UCLA Ace solver, see Chavira & Darwiche, 2008) and the approximate marginal output by the solver. If a solver does not output a solution, we consider its KL-distance to be $\infty$. A score lies between 0 and 1, with 1 indicating that the solver outputs exact solution while 0 indicating that the solver either does not output a solution or has infinite average KL distance. Figure 27 shows the score as a function of KL distance.

In Figures 28-35 we report the results of experiments with each of the problem sets. Each solver has a timeout of 20 minutes on each problem instance; when solving a problem, each solver periodically outputs the best solution found so far. Using this, we can compute, for each solver, at any point in time, the total sum of its scores over all problem instances in a particular set, called SumScore(t). On the horizontal axis, we have the time and on the vertical axis, the SumScore(t). The higher the curve of a solver is, the better (the higher the score).

In summary, we see that IJGP shows the best performance on the first four classes of networks (UAI-MPE, UAI-PE, Relational and Linkage), it is tied with other algorithms on two classes (Grid and BN2O), and is surpassed by EDBP on the last two classes (WCSPs and Promedas). EPIS and SampleSearch, which are importance sampling schemes, are often inferior to IJGP and EDBP. In theory, the accuracy of these importance sampling schemes should improve with time. However, the rate of improvement is often unknown in practice. On the hard benchmarks that we evaluated on, we found that this rate is quite small and therefore the improvement cannot be discerned from the Figures. We discuss the results in detail below.

As mentioned earlier, TLSBP works only on binary networks (i.e., two variables per function) and therefore it was not evaluated on WCSPs, Linkage, UAI06-MPE and UAI06-PE benchmarks.

The UAI-MPE and UAI-PE instances were used in the UAI 2006 evaluation of exact solvers (for details see the report by Bilmes & Dechter, 2006). Exact marginals are available on 40 UAI-MPE instances and 58 UAI-PE instances. The results for UAI-MPE and UAI-PE instances are shown in Figures 28 and 29 respectively. IJGP is the best performing scheme on both benchmark sets reaching a SumScore very close to the maximum possible value in both cases after about 2 minutes of CPU time. EDBP and SampleSearch are second best in both cases.

Relational network instances are generated by grounding the relational Bayesian networks using the Primula tool (Chavira, Darwiche, & Jaeger, 2006). Exact marginals are available only on 30 out of the submitted 251 instances. From Figure 30, we observe that IJGP's SumScore steadily increases with time and reaches a value very close to the maximum possible score of 30 after about 16 minutes of CPU time. SampleSearch is the second best performing scheme. EDBP, TLSBP and EPIS perform quite poorly on these instances reaching the SumScore of 10, 13 and 13 respectively after 20 minutes of CPU time.

The Linkage instances are generated by converting linkage analysis data into a Markov network using the Superlink tool (Fishelson & Geiger, 2003). Exact marginals are available only on 5 out of the 22 instances. The results are shown in Figure 31. After about one minute of CPU time, IJGP's SumScore is close to 5 which remains steady thereafter while EDBP only reaches a SumScore of 2 in 20 minutes. SampleSearch is the second best performing scheme while EDBP is third best.

The results on Grid networks are shown in Figure 32. The sink node of the grid is the evidence node. The deterministic ratio $p$ is a parameter specifying the fraction of nodes that are deterministic, that is, whose values are determined given the values of their parents. The evaluation benchmark set consists of 30 instances having $p = 50\%, 75\%$ and $90\%$ with exact marginals available on 27 instances only. EPIS, IJGP, SampleSearch and EDBP are in a close tie on this network, while TLSBP has the lowest performance. While hard to see, EPIS is just slightly the best performing scheme, IJGP is the second best followed by SampleSearch and EDBP. On this instances IJGP's SumScore increases steadily with time.

The results on BN2O instances appear in Figure 33. This is again a very close tie, in this case of all five algorithms. IJGP has a minuscule decrease of SumScore with time from 17.85 to 17.7. Although in general an improvement in accuracy is expected for IJGP with higher i-bound, it is not

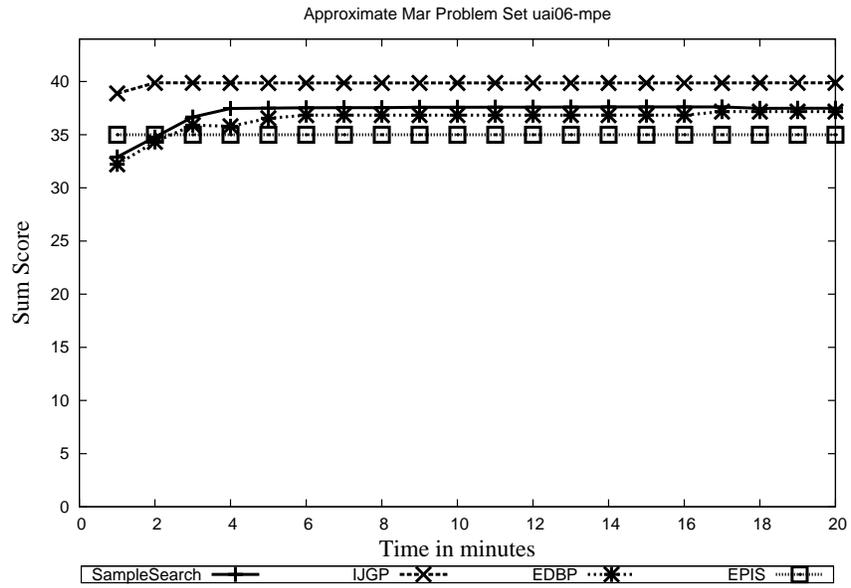

Figure 28: Results on UAI-MPE networks. TLSBP is not plotted because it cannot handle UAI-MPE benchmarks.

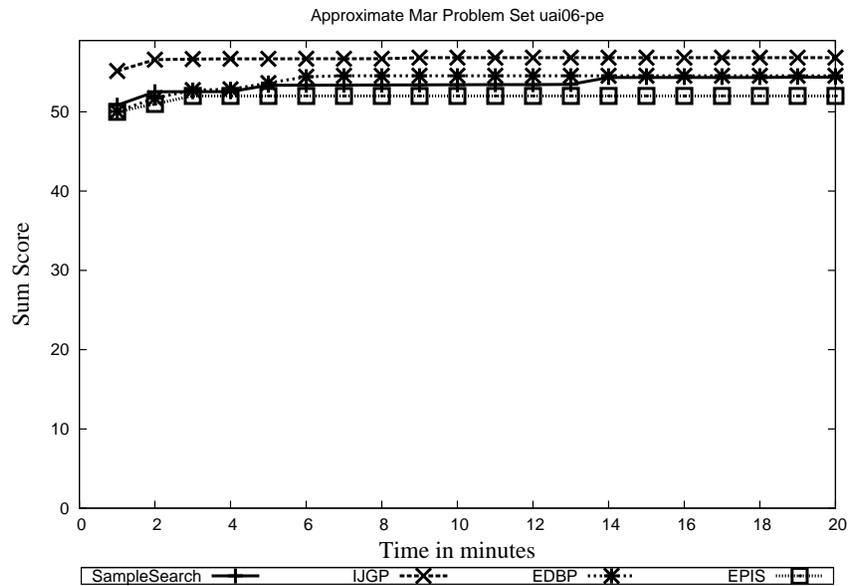

Figure 29: Results on UAI-PE networks. TLSBP is not plotted because it cannot handle UAI-PE benchmarks.

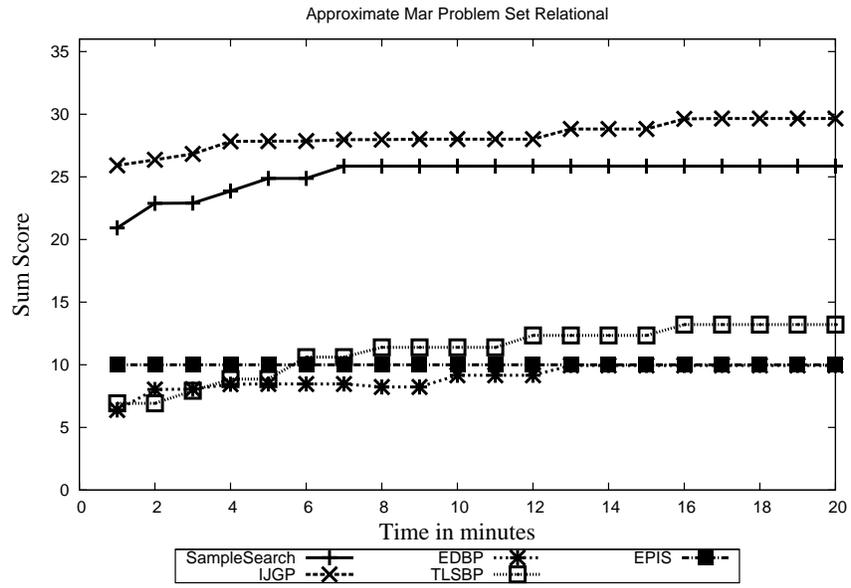

Figure 30: Results on relational networks.

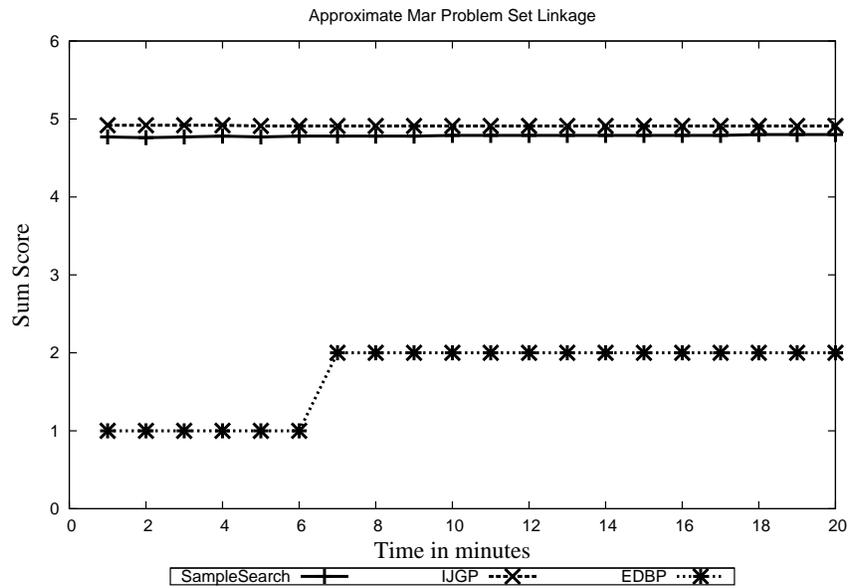

Figure 31: Results on Linkage networks. EPIS and TLSBP are not plotted because they cannot handle Linkage networks.

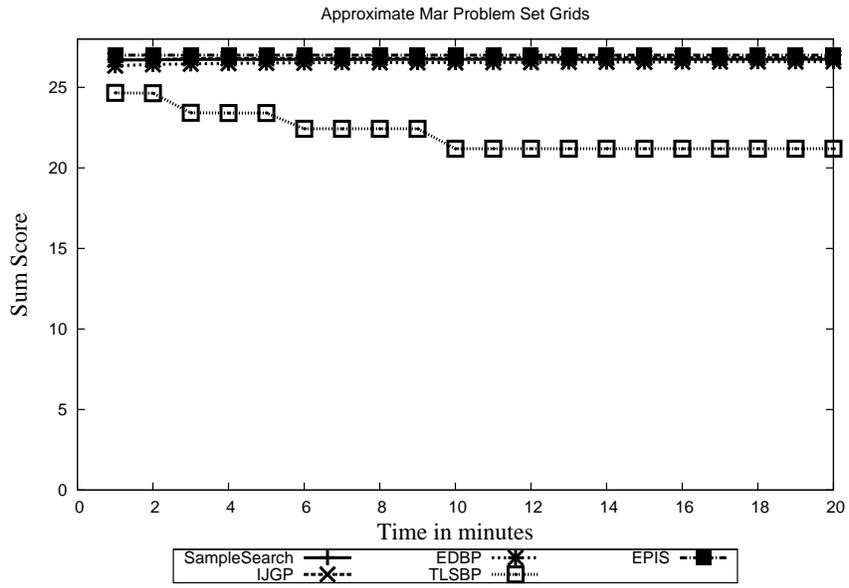

Figure 32: Results on Grid networks.

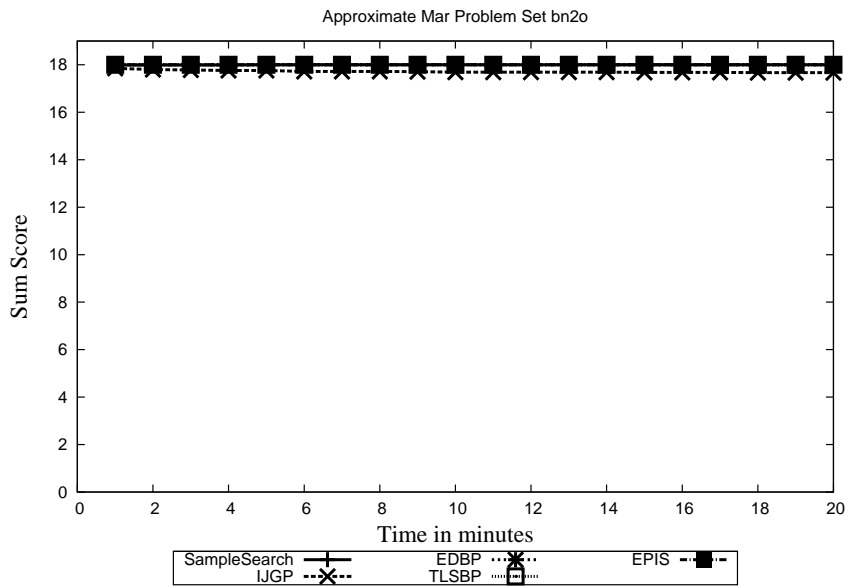

Figure 33: Results on BN2O networks. All solvers except IJGP quickly converge to the maximum possible score of 18 and are therefore indistinguishable in the Figure.

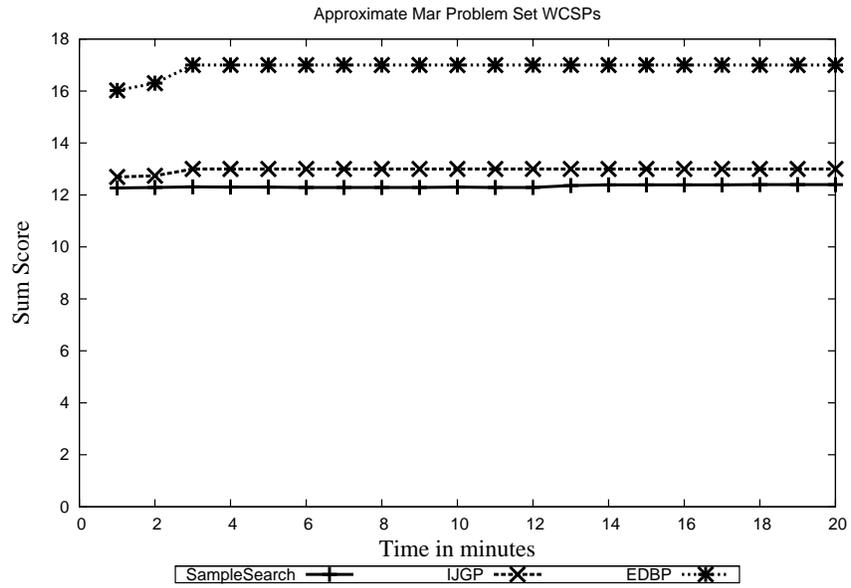

Figure 34: Results on WCSPs networks. EPIS and TLSBP are not plotted because they cannot handle WCSPs.

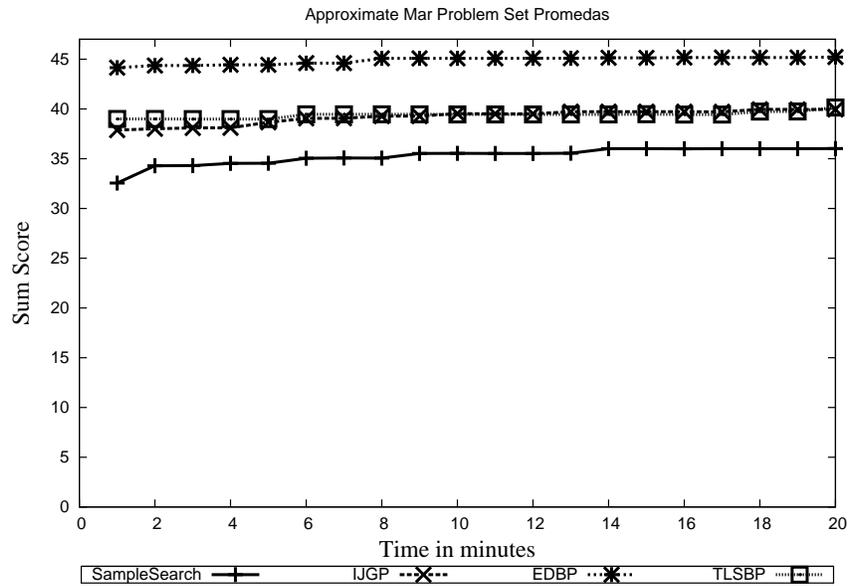

Figure 35: Results on Promedas networks. EPIS is not plotted because it cannot handle Promedas benchmarks, which are Markov networks.

guaranteed, and this is an example when it does not happen. The other solvers reach the maximum possible SumScore of 18 (or very close to it) after about 6 minutes of CPU time.

The WCSP benchmark set has 97 instances. However we used only the 18 instances for which exact marginals are available. Therefore the maximum SumScore that an algorithm can reach is 18. The results are shown in Figure 34. EDBP reaches a SumScore of 17 after almost 3 minutes of CPU time while IJGP reaches a SumScore of 13 after about 3 minutes. The SumScores of both IJGP and EDBP remain unchanged in the interval from 3 to 20 minutes. After looking at the raw results, we found that IJGP's score was zero on 5 instances out of 18. This was because the singleton consistency component implemented via the SAT solver did not finish in 20 minutes on these instances. Although the singleton consistency step generally helps to reduce the practical time complexity of IJGP on most instances, it adversely affects it on these WCSP instances.

The Promedas instances are Noisy-OR binary Bayesian networks (Pearl, 1988). These instances are characterized by extreme marginals. Namely, for a given variable, the marginals are of the form $(1 - \epsilon, \epsilon)$ where $\epsilon$ is a very small positive constant. Exact marginals are available only on 46 out of the submitted 238 instances. On these structured problems (see Figure 35), we see that EDBP is the best performing scheme reaching a SumScore very close to 46 after about 7 minutes of CPU time while TLSBP and IJGP are able to reach a SumScore of about 40 in 20 minutes.

## 6. Related Work

There are numerous lines of research devoted to the study of belief propagation algorithms, or message-passing schemes in general. Throughout the paper we have mentioned and compared with other related work, especially in the experimental evaluation section. We give here a short summary of the developments in belief propagation and present some related schemes that were not mentioned before. For additional information see also the recent review by Koller (2010).

About a decade ago, Iterative Belief Propagation (Pearl, 1988) received a lot of interest from the information theory and coding community. It was realized that two of the best error-correcting decoding algorithms were actually performing belief propagation in networks with cycles. The LDPC code (low-density parity-check) introduced long time ago by Gallager (1963), is now considered one of the most powerful and promising schemes that often performs impressively close to Shannon's limit. Turbo codes (Berrou, Glavieux, & Thitimajshima, 1993) are also very efficient in practice and can be understood as an instance of belief propagation (McEliece et al., 1998).

A considerable progress towards understanding the behavior and performance of BP was made through concepts from statistical physics. Yedidia et al. (2001) showed that IBP is strongly related to the Bethe-Peierls approximation of variational (Gibbs) free energy in factor graphs. The Bethe approximation is a particular case of the more general Kikuchi (1951) approximation. Generalized Belief Propagation (Yedidia et al., 2005) is an application of the Kikuchi approximation that works with clusters of variables, on structures called region graphs. Another algorithm that employs the region-based approach is Cluster Variation Method (CVM) (Pelizzola, 2005). These algorithms focus on selecting a good region-graph structure to account for the over-counting (and over-over-counting, etc.) of evidence. We view generalized belief propagation more broadly as any belief propagation over nodes which are clusters of functions. Within this view IJGP, and GBP as defined by Yedidia et al. (2001), as well as CVM, are special realizations of generalized belief propagation.

Belief Propagation on Partially Ordered Sets (PBP) (McEliece & Yildirim, 2002) is also a generalized form of Belief Propagation that minimizes the Bethe-Kikuchi variational free energy, and that works as a message-passing algorithm on data structures called partially ordered sets, which has junction graphs and factor graphs as examples. There is one-to-one correspondence between fixed points of PBP and stationary points of the free energy. PBP includes as special cases many other variants of belief propagation. As we noted before, IJGP is basically the same as PBP.

Expectation Propagation (EP) (Minka, 2001) is a an iterative approximation algorithm for computing posterior belief in Bayesian networks. It combines assumed-density filtering (ADF), an extension of the Kalman filter (used to approximate belief states using expectations, such as mean and variance), with IBP, and iterates until these expectations are consistent throughout the network. TreeEP (Minka & Qi, 2004) deals with cyclic problem by reducing the problem graph to a tree subgraph and approximating the remaining edges. The relationship between EP and GBP is discussed by Welling, Minka, and Teh (2005).

Survey Propagation (SP) (Braunstein et al., 2005) solves hard satisfiable (SAT) problems using a message-passing algorithm on a factor graph consisting of variable and clause nodes. SP is inspired by an algorithm called Warning Propagation (WP) and by BP. WP can determine if a tree-problem is SAT, and if it is then it can provide a solution. BP can compute the number of satisfying assignments for a tree-problem, as well as the fraction of the assignments where a variable is true. These two algorithms are used as heuristics to define the SP algorithm, that is shown to be more efficient than either of them on arbitrary networks. SP is still a heuristic algorithm with no guarantee of convergence. SP was inspired by the new concept of "cavity method" in statistical physics, and can be interpreted as BP where variables can not only take the values true or false, but also the extra "don't care" value. For a more detailed treatment see the book by Mézard and Montanari (2009).

## 7. Conclusion

In this paper we investigated a family of approximation algorithms for Bayesian networks, that could also be extended to general graphical models. We started with bounded inference algorithms and proposed Mini-Clustering (MC) scheme as a generalization of Mini-Buckets to arbitrary tree decompositions. Its power lies in being an anytime algorithm governed by a user adjustable i-bound parameter. MC can start with small i-bound and keep increasing it as long as it is given more time, and its accuracy usually improves with more time. If enough time is given to it, it is guaranteed to become exact. One of its virtues is that it can also produce upper and lower bounds, a route not explored in this paper.

Inspired by the success of iterative belief propagation (IBP), we extended MC into an iterative message-passing algorithm called Iterative Join-Graph Propagation (IJGP). IJGP operates on general join-graphs that can contain cycles, but it is sill governed by an i-bound parameter. Unlike IBP, IJGP is guaranteed to become exact if given enough time.

We also make connections with well understood consistency enforcing algorithms for constraint satisfaction, giving strong support for iterating messages, and giving insight into the performance of IJGP (IBP). We show that: (1) if a value of a variable is assessed as having zero-belief in any iteration of IJGP, then it remains a zero-belief in all subsequent iterations; (2) IJGP converges in a finite number of iterations relative to its set of zero-beliefs; and, most importantly (3) that the set of zero-beliefs decided by any of the iterative belief propagation methods is sound. Namely any zero-belief determined by IJGP corresponds to a true zero conditional probability relative to the given probability distribution expressed by the Bayesian network.

Our experimental evaluation of IJGP, IBP and MC is provided, and IJGP emerges as one of the most powerful approximate algorithms for belief updating in Bayesian networks.